\documentclass[journal]{IEEEtran}

\usepackage{fontawesome5}

\usepackage{amsmath}
\usepackage{amssymb}
\usepackage{amsthm}
\usepackage[table,dvipsnames]{xcolor}
\usepackage{booktabs}
\usepackage{multirow}
\usepackage{etoolbox}
\usepackage{siunitx}
\usepackage{microtype}
\usepackage[most]{tcolorbox}

\usepackage[hyphens]{url}
\usepackage{hyperref}
\hypersetup{%
  hidelinks,%
  colorlinks=true,%
  raiselinks=true,%
  allcolors=black,%
  pdfstartview=Fit,%
  breaklinks=true,%
  hypertexnames=false,
}
\usepackage{xr}
\usepackage[capitalise,nameinlink]{cleveref}

\usepackage[all]{hypcap}
\usepackage{graphicx}
\usepackage[caption=false,font=normalsize,labelfont=sf,textfont=sf]{subfig}
\usepackage{textcomp}
\usepackage{stfloats}
\usepackage{cite}
\usepackage{balance}

\usepackage{algorithm}
\usepackage{algpseudocode}

\usepackage{pdfpages} 
\AtEndDocument{%
  \clearpage
  \includepdf[pages=-,]{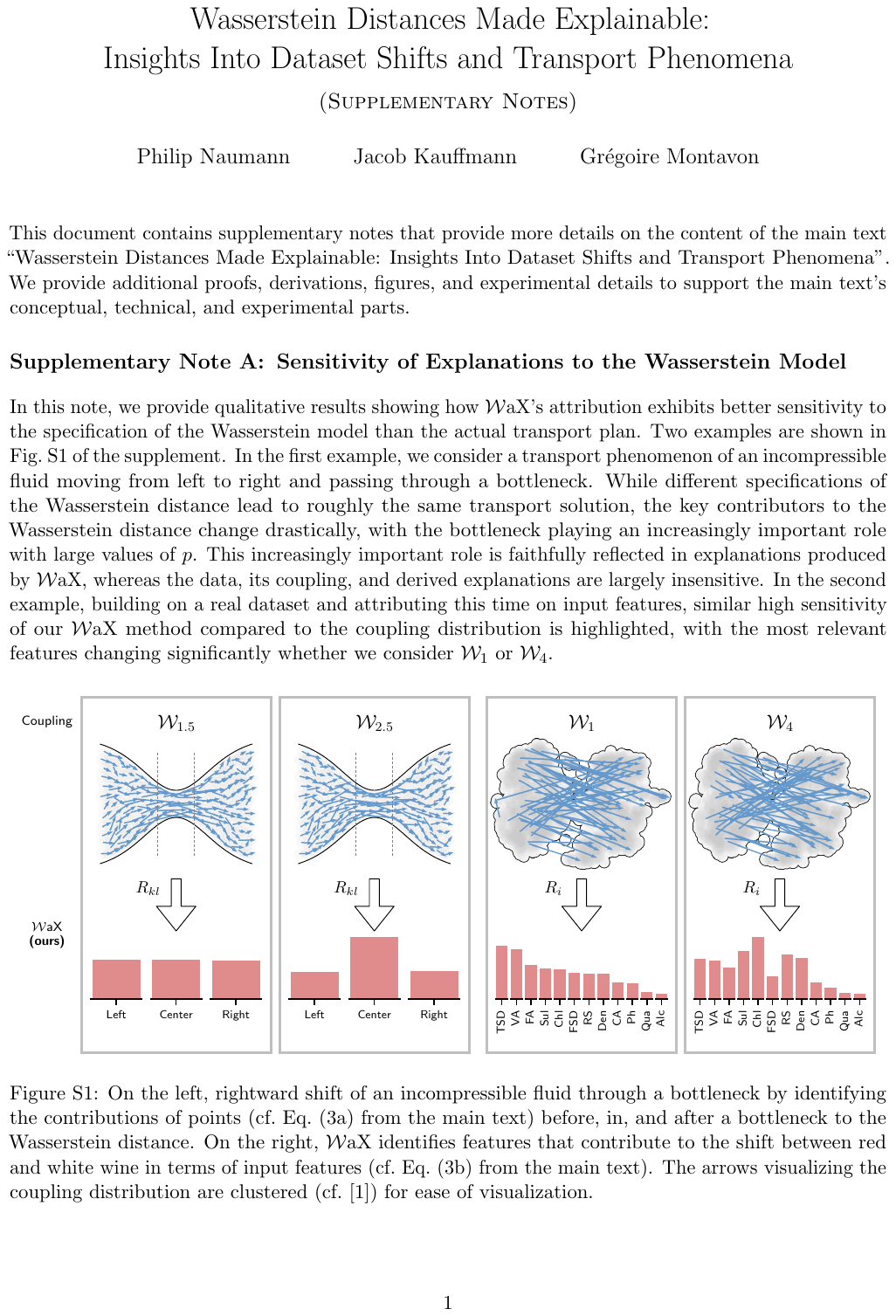}%
}%

\newcommand{\E}{\mathbb{E}}
\newcommand{\bone}{\boldsymbol{1}}

\newtcolorbox{fwbox}{
    enhanced,
    breakable = true,
    boxrule=0.5pt,
    boxsep=0pt,
    top=-3pt,
    lines before break=5,
    width=.9\linewidth,
    colback = white,
    colframe = white,
    borderline={0.5mm}{0mm}{Gray!60!white}
    }

\newtcolorbox{lrpbox}{
    enhanced,
    breakable = true,
    boxrule=0.5pt,
    boxsep=0pt,
    top=-3pt,
    lines before break=5,
    width=.9\linewidth,
    colback = white,
    colframe = white,
    borderline={0.5mm}{0mm}{WildStrawberry!40!white}
    }

\newtcolorbox{lrpquickbox}{
    enhanced,
    breakable = true,
    boxrule=0.5pt,
    boxsep=0pt,
    top=-3pt,
    lines before break=5,
    width=.9\linewidth,
    colback = white,
    colframe = white,
    borderline={0.5mm}{0mm}{WildStrawberry!40!white,dotted}
    }

\newcommand{\supp}[1]{Supplementary Note~#1} 
\newcommand{\mname}{$\mathcal{W}$aX}
\newcommand{\wdsa}{$\boldsymbol{U}$-\mname{}}

\newtheorem{proposition}{Proposition}

\title{Wasserstein Distances Made Explainable: Insights Into Dataset Shifts and Transport Phenomena}

\author{Philip Naumann, Jacob Kauffmann, Gr\'egoire Montavon
\thanks{P.~Naumann is with
BIFOLD\,--\,Berlin Institute for the Foundations of Learning and Data, 10587 Berlin, Germany,
and
the Department of Electrical Engineering and Computer Science, Technische Universit\"at Berlin, 10587 Berlin, Germany
}
\thanks{J.~Kauffmann is with
BIFOLD\,--\,Berlin Institute for the Foundations of Learning and Data, 10587 Berlin, Germany,
and
the Department of Electrical Engineering and Computer Science, Technische Universit\"at Berlin, 10587 Berlin, Germany
}
\thanks{G.~Montavon is with BIFOLD\,--\,Berlin Institute for the Foundations of Learning and Data, 10587 Berlin, Germany,
and the
Institute for AI in Medicine, Charit\'e\,--\,Universit\"atsmedizin Berlin, 10117 Berlin, Germany.
E-mail: gregoire.montavon@charite.de.
}
\thanks{(Corresponding Author: Gr\'egoire Montavon)}
}

\begin{document}
\robustify\bfseries
\robustify\itshape

\maketitle

\begin{abstract}
Wasserstein distances provide a powerful framework for comparing data distributions. They can be used to analyze processes over time or to detect inhomogeneities within data. However, simply calculating the Wasserstein distance or analyzing the corresponding transport plan (or coupling) may not be sufficient for understanding what factors contribute to a high or low Wasserstein distance. In this work, we propose a novel solution based on Explainable AI that allows us to efficiently and accurately attribute Wasserstein distances to various data components, including data subgroups, input features, or interpretable subspaces. Our method achieves high accuracy across diverse datasets and Wasserstein distance specifications, and its practical utility is demonstrated in three use cases.
\end{abstract}


\section{Introduction}
\label{sec:intro}

\IEEEPARstart{I}{n many} fields of research, it is necessary to analyze data at the population level.
For example, when analyzing chemical reactions or transport phenomena, one typically cannot track every single atom of the physical system of interest and must rely on densities instead.
Similarly, when trying to characterize a medical disease, one may not have access to each patient in multiple health states, but only to non-overlapping cohorts of healthy and unhealthy patients.
As another example, complex processes that unfold over time (e.g.~the spread of an epidemic or the development of biological cells~\cite{kleinMappingCellsTime2025}) may only be analyzable at discrete points in time without the ability to trace their detailed temporal evolution.

\smallskip

The Wasserstein distance is a popular and theoretically sound tool for making comparisons at the distribution level.
It has its roots in Optimal Transport (OT)~\cite{villaniOptimalTransportOld2008,peyreComputationalOptimalTransport2020,montesumaRecentAdvancesOptimal2025}, a field concerned with finding cost-minimizing couplings between source and target distributions.
The Wasserstein distance can be defined as the result of the optimization problem:
\begin{align}
\mathcal{W}_p(\mu,\nu) &= \inf_{\gamma \in \Gamma(\mu,\nu)} \big(\mathbb{E}_{(x,y) \sim \gamma} \|x-y\|_q^p \big)^{1/p} 
\label{eq:wasserstein}
\end{align}
s.t.~$\gamma \bone = \mu, \; \gamma^\top \bone = \nu, \; \gamma \ge 0$, where $\mu$ and $\nu$ are the two data distributions, $\Gamma(\mu,\nu)$ is the set of all possible couplings $\gamma$ between them, and $p \geq 1$ is a hyperparameter that emphasizes penalizing large distances.
The minimization objective in \cref{eq:wasserstein} can be interpreted as applying a least-action principle in determining how to match points from the two distributions.
Moreover, regularized formulations such as Sinkhorn~\cite{cuturiSinkhornDistancesLightspeed2013} are also common in practice.

\smallskip

When interested specifically in understanding the nature of the distribution shift, looking at the Wasserstein distance itself provides only limited insight. It only measures the extent to which the distributions $\mu$ and $\nu$ differ without highlighting the relevant input features or data points.
The coupling (or transport plan) $\gamma^\star$ obtained as a byproduct of calculating the Wasserstein distance using \cref{eq:wasserstein} can provide further insight into the dataset shift (e.g.~\cite{kulinskiExplainingDistributionShifts2023}). Yet, it still does not specifically pinpoint which part of the data or which input features are genuinely responsible for the low or high Wasserstein distance (cf.~\supp{A}).

\smallskip

In this work, we address the question of what in the data contributes to the Wasserstein distance between distributions using the tools of Explainable AI (XAI). Explainable AI~\cite{montavonMethodsInterpretingUnderstanding2018,gunningDARPAsExplainableArtificial2019,samekExplainingDeepNeural2021,barredoarrietaExplainableArtificialIntelligence2020} is a growing subfield of Machine Learning (ML) that aims to reveal the complex inner-workings of black-box ML models, whether it is to detect flaws in their decision strategy~\cite{lapuschkinUnmaskingCleverHans2019,degraveAIRadiographicCOVID192021,kauffmannExplainableAIReveals2025}, or provide novel insights~\cite{binderMorphologicalMolecularBreast2021,klauschenExplainableArtificialIntelligence2024,zednikScientificExplorationExplainable2022,roscherExplainableMachineLearning2020}. Methods such as Shapley values~\cite{strumbeljExplainingPredictionModels2014,lundbergUnifiedApproachInterpreting2017} or Layer-wise Relevance Propagation (LRP)~\cite{bachPixelwiseExplanationsNonlinear2015,montavonLayerWiseRelevancePropagation2019}, have been successful in attributing the predictions of a wide variety of popular ML models in terms of input features. To the best of our knowledge, there have been no systematic studies on how to explain the distance between data, especially when distances are not considered at the single-instance level, but at the level of a whole distribution (e.g.~the Wasserstein distance).

\smallskip

\begin{figure*}[t!]
    \centering
    \includegraphics[width=\linewidth]{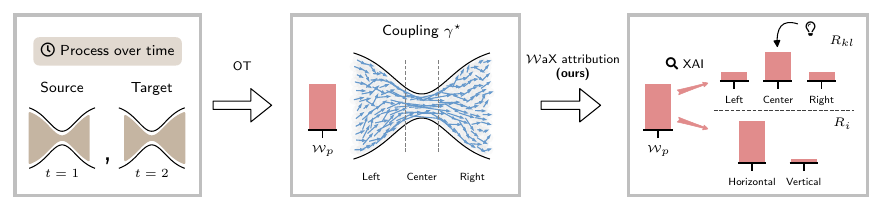}
    \caption{
    High-level illustration of \mname{}, which goes beyond a classical inspection of the Wasserstein-derived transport plan $\gamma^\star$, by attributing the Wasserstein distance on data points and input features. In this example, a fluid moves rightward and is observed at times $t=1$ and $t=2$. This phenomenon is represented by a Wasserstein model ($p=3$, $q=2$). \mname{} identifies the bottleneck at the center as the primary contributor to the high Wasserstein distance---something that is not easily detectable and quantifiable from analyzing the transport plan alone. In \supp{A}, we provide further examples highlighting how using \mname{} and simply analyzing the transport plan strongly differ in their sensitivity to different Wasserstein model specifications.
    }
    \label{fig:explaining-wasserstein-vs-coupling}
\end{figure*}

We propose a new set of Explainable AI techniques referred to together as ``\mname{}'' (or `Wasserstein distances made explainable') focused on attributing the Wasserstein distance in terms of various aspects of the data, such as individual data points or input features (see \cref{fig:explaining-wasserstein-vs-coupling,fig:wax-attribution-schematic,sec:xai-ot}).
Our proposed method is grounded in the framework of LRP, expressing a precomputed Wasserstein distance model as a neural network and explaining it by reverse propagating in that graph. The proposed approach satisfies standard axiomatic properties of explanations, such as conservation, and demonstrates broad applicability by supporting both classical Wasserstein distances and their Sinkhorn-regularized variants~\cite{cuturiSinkhornDistancesLightspeed2013}. Additionally, it accommodates instance-wise comparisons using the general class of Minkowski distances, which includes the Euclidean distance as a special case. Our method also integrates with subspace analysis methods (see \cref{sec:concept-based-xot}), thereby instantiating ``Concept-based XAI''~\cite{kimInterpretabilityFeatureAttribution2018,achtibatAttributionMapsHumanunderstandable2023,chormaiDisentangledExplanationsNeural2024} to the problem of explaining Wasserstein distances.

\smallskip

Through quantitative evaluations on several datasets and Wasserstein distance specifications, we show that our method consistently outperforms a representative set of baseline methods (see \cref{sec:evaluation}). The practical utility of our method is demonstrated in three use cases (see \cref{sec:use-case-relevant-feature-discovery,sec:use-case-abalone-age,sec:use-case-dataset-differences}), where it aids in making ML classifiers more robust, provides insight into complex, nonlinear transport phenomena, and helps to navigate differences between large multimodal image datasets.

Furthermore, \mname{} is \emph{model-centric}: it analyzes data through the lens of a specific Wasserstein distance model, thereby giving users control over how the data is viewed. For instance, one can choose the Wasserstein formulation itself (e.g.~Sinkhorn~\cite{cuturiSinkhornDistancesLightspeed2013} versus exact optimal transport), select the Wasserstein power $p$ to tune sensitivity to outliers (with larger $p$ increasing sensitivity), and adjust the underlying distance metric via the parameter $q$ to emphasize particular geometric aspects of the data.
Overall, our method offers capabilities that go well beyond classical statistical analysis or manual inspection of the data.

\section{Related Work}
\label{sec:related-work}

\subsubsection*{Causal Models of Distribution Shifts} Dealing with distribution shifts is a well-known problem in machine learning~\cite{quionero-candelaDatasetShiftMachine2009, kohWILDSBenchmarkIntheWild2021}. Substantial research has focused on better handling (e.g.~\cite{courtyOptimalTransportDomain2017}) and understanding these shifts. One approach involves characterizing distribution shifts through the lens of causal relationships specified via conditional distributions~\cite{budhathokiWhyDidDistribution2021}. Within this framework, the Kullback-Leibler Divergence (KLD) can be used to quantify the magnitude of the shift, and Shapley values help to attribute this divergence back to input variables. Another recent work by Babbar {\em et al.}~\cite{babbarWhatDifferentThese2025} focuses on better understanding what makes two datasets different by proposing a framework for explaining distribution shifts. For this purpose, they use `Rashomon Importance Distributions'~\cite{donnellyRashomonImportanceDistribution2023}, which are created by sampling importance scores from a set of well-performing (ML) surrogate models. In contrast, our work uses the Wasserstein distance as a more \textit{direct} distribution shift model from which a shift can be more directly characterized in comparison to, e.g., linear models like logistic regression classifiers, as demonstrated in \cref{sec:external-evaluation}.

\subsubsection*{Interpreting OT Maps} Another related line of work uses the OT map to retrieve information about the modeled shift. Kulinski {\em et al.}~\cite{kulinskiExplainingDistributionShifts2023} propose two methods for this purpose: $k$-sparse and $k$-cluster maps.
The former expresses the $k$ most relevant shift features as a new transport map, which they find by looking at the mean shift strengths between the distributions and sorting them according to their magnitude.
The $k$-cluster maps model heterogeneous shifts using $k$-means clustering on the column-wise concatenated source and transport vectors. They aim to detect sub-shifts, which are then expressed through individual transport maps. Another related work~\cite{hulkundInterpretableDistributionShift2022} also uses the OT map to interpret distribution shifts in classification datasets by comparing distances of farthest and closest examples. 
Finally, Gautheron {\em et al.}~\cite{gautheronFeatureSelectionUnsupervised2018} propose using OT between the features of previously aligned domains to detect and prune dissimilar dimensions between them, thereby improving the efficiency of domain adaptation.
Unlike the above works, our approach does not explain distribution shifts based on the Wasserstein-derived transport map or between the features, but directly based on the Wasserstein \textit{distance} (cf.~\cref{fig:explaining-wasserstein-vs-coupling}). This enables us to frame the explanation task as an attribution problem (cf.~\cite{strumbeljExplainingPredictionModels2014, sundararajanAxiomaticAttributionDeep2017, bachPixelwiseExplanationsNonlinear2015}), and derive relevance scores that directly capture the effect of adding or removing input features on the distance between distributions (see \cref{sec:internal-evaluation}).

\subsubsection*{Intrinsically Interpretable OT Models} Some works have focused on making OT itself more interpretable through specific reformulations of the OT problem that include anchors or feature sparsity~\cite{linMakingTransportMore2021,cuturiMongeBregmanOccam2023}. In contrast to these works, we propose a post-hoc explanation method that can readily apply to popular Wasserstein distance formulations. In particular, our method can explain transport phenomena without introducing specific modeling constraints.

\subsubsection*{Explaining Robustness under Distribution Shift} Recent advances focus on explaining the behavior of (classification) ML models under distribution shifts by incorporating data relationships. Zhang {\em et al.}~\cite{zhangWhyDidModel2023} extend the work of~\cite{budhathokiWhyDidDistribution2021} by focusing on explaining the performance drop of an ML model due to an underlying distribution shift.
Using a causal graph and Shapley values, they trace the identified causes back to the data distributions. Recent work used the OT coupling to find features likely to affect a classifier's performance~\cite{deckerExplanatoryModelMonitoring2024}.
They also use Shapley values to assess the contribution of each feature to the shift-induced performance loss. Unlike these works, our approach does not explain the behavior of an ML model under distribution shifts, but focuses on characterizing the shift itself. In the context of tabular data, Liu {\em et al.}~\cite{liuNeedLanguageDescribing2023} argue that not all distribution shifts are equal.
Different variants, stemming from various causes, each require special attention.
For this reason, they propose a benchmark that focuses on ``$(Y | X)$-shifts'', which they consider the most critical in affecting ML performance. Complementary to~\cite{liuNeedLanguageDescribing2023}, another benchmark has been proposed by Gardner {\em et al.}~\cite{gardnerBenchmarkingDistributionShift2023}.
Our work differs in its goals by attempting to characterize all types of shifts, and methodologically by identifying each feature's contribution.

\subsubsection*{Explaining Distance-Based Models} Several papers have addressed the technical question of explaining distances in ML models. Specific instantiations have been proposed in the context of anomaly~\cite{montavonExplainingPredictionsUnsupervised2022} and clustering models~\cite{kauffmannClusteringClusterExplanations2024}, both of which study the particular geometry of the distance function and the composition of these in the prediction model. Our work differs in that we consider distances not only at the instance level, but also at the level of the entire data distribution.

\subsubsection*{OT as a Tool for Explainability} The intersection of Explainable AI and optimal transport has been explored for broader goals. These include generating counterfactual explanations at the distribution level~\cite{youDistributionalCounterfactualExplanations2025}, making ML models more interpretable~\cite{serrurierExplainableProperties1Lipschitz2023}, and evaluating XAI methods themselves~\cite{hillAxiomaticExplainerGlobalness2025}. In all cases, the goals of these works diverge significantly from our main goal, which is to characterize distribution shifts and transport phenomena.

\section{Wasserstein Distances Made Explainable}
\label{sec:xai-ot}

We now introduce our proposed method, called ``\mname{}'', for explaining Wasserstein distances in terms of data points and input features. \mname{} applies to a fairly general class of Wasserstein distances ($\mathcal{W}_p$ with arbitrary exponent $p$ and the class of Minkowski distances between pairs of points). It also allows the user to flexibly choose the basis for interpretability, as it can produce an attribution in terms of individual instances, input features, latent subspaces (cf.~\cref{sec:concept-based-xot}), or combinations of these. Technically, \mname{} follows a \textit{neuralization-propagation} approach (cf.~\cite{kauffmannClusteringClusterExplanations2024}), where the quantity to be explained, here the Wasserstein distance, is first rewritten as a functionally equivalent neural network, which then guides the explanation process through a purposefully designed backward pass following propagation rules inspired by LRP\footnote{Explainable AI methods such as LRP~\cite{bachPixelwiseExplanationsNonlinear2015}, guided GradCAM~\cite{selvarajuGradCAMVisualExplanations2020}, or DeepSHAP~\cite{lundbergUnifiedApproachInterpreting2017} all require the ML model to have a neural network structure, and they explain by performing a backward pass in this neural network.}.

\textit{Layer-wise Relevance Propagation (LRP)} is a widely used technique for explaining the predictions of machine learning models, in particular, large neural networks~\cite{bachPixelwiseExplanationsNonlinear2015,aliXAITransformersBetter2022}, and has extensions to other model classes (e.g.~\cite{kauffmannClusteringClusterExplanations2024,schnakeHigherorderExplanationsGraph2022}).
Conceptually, LRP views a trained model as a computational graph and assigns a relevance score to each input feature by propagating the model's output backward through this graph. This backward pass is governed by a set of `LRP rules' tailored to the individual layers or components of the model, with the aim of producing explanations that faithfully explain the model's output at a low computational cost.
\textit{Neuralization}, on the other hand, aims to bring the benefits of LRP to a broader set of architectures, including distance- and kernel-based models~\cite{kauffmannClusteringClusterExplanations2024}. It attempts, through algebraic manipulations, to rewrite the original model as a sequence of layers that are `neural-network-like' (e.g.~interpretable as detection and pooling layers). Neuralization fundamentally differs from more classical surrogate modeling approaches by \textit{not} involving a training step. This ensures a tight equivalence between the original model to explain and the resulting neural network. 

\subsection{Neuralization-Propagation Procedure}
\label{section:neuralization-propagation}

\begin{figure*}[t!]
\centering
\includegraphics[width=.98\textwidth]{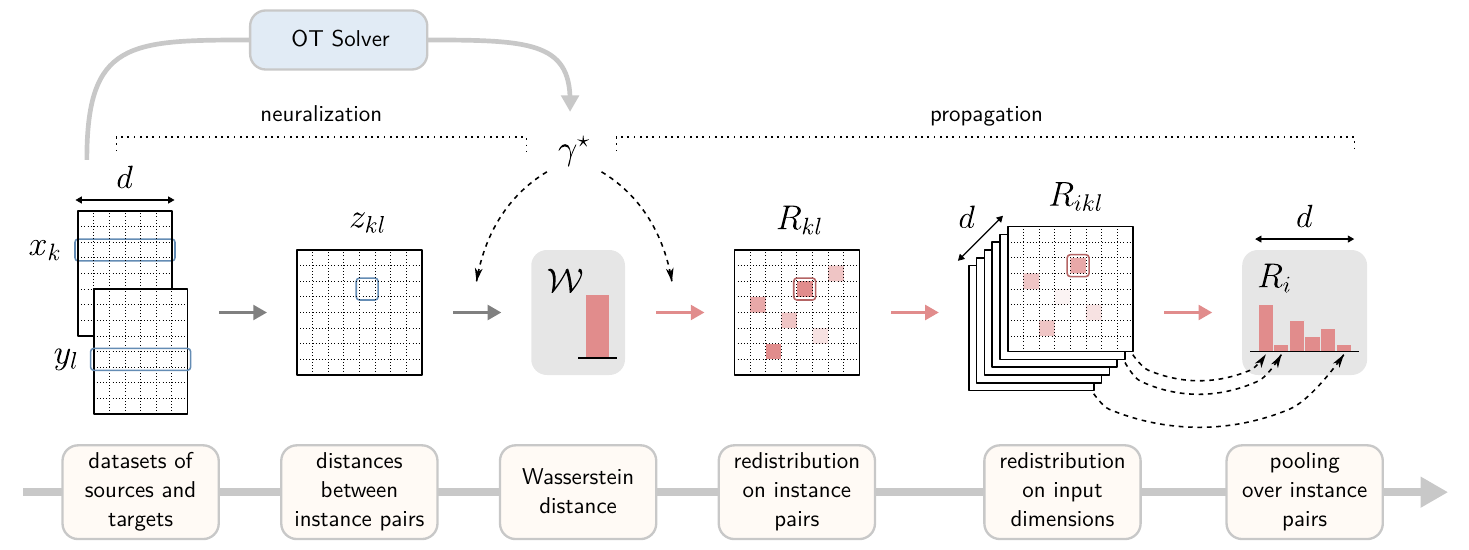}
\caption{Diagram of \mname{}'s workflow. The Wasserstein distance is rewritten as a two-layer network (\cref{eq:fw-1,eq:fw-2}) with the coupling fixed to its optimum $\gamma^\star$. Explanation proceeds through a backward pass, leading first to an attribution on instance pairs (\cref{eq:wax-attribution-sample}), and then on input features (\cref{eq:wax-attribution-feature}).}
\label{fig:wax-attribution-schematic}
\end{figure*}

The \mname{} approach (illustrated in \cref{fig:wax-attribution-schematic}) instantiates the neuralization-propagation framework described above to Wasserstein distances. In the following, we assume that the distributions $\mu$ and $\nu$ are empirical, i.e.~formally defined as $\mu(\xi) = \frac1N \sum_{k=1}^N \delta(\xi - x_k)$, $\nu(\upsilon) = \frac1M \sum_{l=1}^M \delta(\upsilon - y_l)$, where $\delta$ is the Dirac delta function, and $x_k \in \mathbb{R}^d$ and $y_l \in \mathbb{R}^d$ are data points. The coupling distribution $\gamma$ can then be represented as a matrix of size $N \times M$.

\medskip

\paragraph*{Neuralization} The calculation of the Wasserstein distance $\mathcal{W}_p$, as outputted in \cref{eq:wasserstein}, is a complex iterative procedure. To enable neuralization, we propose simplifying the functional relationship between input distributions and the Wasserstein distance by fixing the coupling matrix $\gamma$ to its optimum $\gamma^\star$ and analyzing only the partial dependencies of $\mathcal{W}_p$ to the input distributions. Building on this simplified functional form, we can identify a neural network structure that reproduces the Wasserstein distance $\mathcal{W}_p$, and decomposes it into two layers of computation:
\begin{center}
\begin{fwbox}
\begin{subequations}
\label{eq:wd-graph}
\begin{align}
z_{kl} &= \|x_k-y_l\|_q \label{eq:fw-1}\\
\mathcal{W}_p &= \Big(\sum_{kl} \gamma_{kl}^\star \cdot z_{kl}^p \Big)^{1/p} \label{eq:fw-2}
\end{align}
\end{subequations}
\end{fwbox}
\end{center}
The first layer computes the differences between pairs of instances from the two distributions. The second layer applies a norm (weighted by the coupling $\gamma^\star$, precomputed by solving the optimal transport problem) over the pairwise distances computed in the previous layer. The difference vectors $(x_k-y_l)$ provide the directionality in input space necessary to identify the relevant input features. Similarly, the norm over pairwise distances weighted by the coupling probabilities $\gamma^\star_{kl}$ helps to identify relevant instances. We note that the choice of the \textit{primal} formulation of the Wasserstein distance (rather than its dual formulation) as a starting point for our method is crucial for explanation purposes: distances between instance pairs appear explicitly in the primal, and it is supported by a coupling matrix $\gamma^\star$ that is instrumental in separating the few relevant instance pairs from the many irrelevant ones.\footnote{In comparison to the primal, the input directionality of the Wasserstein \textit{dual}'s Kantorovich potentials is more difficult to capture. The joint distribution $\mu \times \nu$ over which these potentials are integrated does not easily allow for separating relevant from irrelevant instance pairs.}

\medskip

\paragraph*{Propagation} Leveraging the neural network structure defined above, we can now define LRP rules for this structure to explain the original Wasserstein distance model. Our method proceeds by reverse-propagating the output $\mathcal{W}_p$ through the layers to identify the relevant instances and input features. The first propagation step decomposes $\mathcal{W}_p$ into contributions of instance pairs (denoted by $R_{kl}$). The second step further propagates these contributions to identify the feature contributions $R_i$. Specifically, our \mname{} method defines the following LRP rules:
\begin{center}
\begin{lrpbox}
\begin{subequations}
\label{eq:wax}
\begin{align}
R_{kl} &= \frac{\gamma^\star_{kl} \cdot z_{kl}^\alpha}{\sum_{kl} \gamma^\star_{kl} \cdot z_{kl}^\alpha} \mathcal{W}_p \label{eq:wax-attribution-sample}\\
R_{i} &= \sum_{kl}\frac{|x_{ki} - y_{li}|^\beta}{\sum_i |x_{ki} - y_{li}|^\beta} R_{kl}
\label{eq:wax-attribution-feature}
\end{align}
\end{subequations}
\end{lrpbox}
\end{center}
where $\alpha$ and $\beta$ are two hyperparameters to be set. We place our attention on how to set these hyperparameters. Intuitively, $\alpha$ controls the spread of relevance onto the samples, whereas $\beta$ controls the spread onto the features. Lower values in both cases distribute it more evenly, and higher values condense it, resulting in a few samples/features receiving most of the relevance. The special case of setting $\alpha = p$ and $\beta = q$ is especially interesting, as it lets the numerators of \cref{eq:fw-2,eq:wax-attribution-sample} coincide with the terms in \cref{eq:fw-1,eq:wax-attribution-feature}). This instantiation of $\alpha$ and $\beta$, however, has limitations for large values of $q$ where a corresponding increase in $\beta$ tends to produce too localized explanations. We settle on the heuristic $\alpha = p$ and $\beta = \min(p+2, q)$, assuming $p \ll \infty$, and use it in all our evaluations. A systematic analysis of different $\alpha,\beta$ combinations for the `Mice' dataset is given in \supp{B} and supports this heuristic.

The whole \mname{} attribution procedure is recapitulated in \cref{alg:wax_summary}, emphasizing the prerequisite search for an optimal coupling matrix $\gamma^\star$, the identification of a neural network equivalent, and the forward/backward pass on the network to generate explanations.
\begin{algorithm}[h!]
\caption{\mname{} Feature Attribution}
\label{alg:wax_summary}
\begin{algorithmic}
\Require Source $X$, Target $Y$, parameters $p, q, \alpha, \beta$

\Statex \textbf{Step 1: Optimal Transport}
\State Find $\gamma^\star$ by solving \cref{eq:wasserstein}

\Statex \textbf{Step 2: Neuralization}
\State Construct the neural network as defined in \cref{eq:fw-1,eq:fw-2}

\Statex \textbf{Step 3: Explanation}
\State $\mathcal{W}_p \gets$ Forward propagation applying \cref{eq:fw-1,eq:fw-2}
\State $R_i \gets$ Relevance propagation applying \cref{eq:wax-attribution-sample,eq:wax-attribution-feature}

\State \Return $R_i$
\end{algorithmic}
\end{algorithm}%

\subsection{Theoretical Properties}

On a more theoretical note, \mname{} fulfills a number of properties. A first property is \textit{conservation}: It is easy to see from the above equations that the LRP rules satisfy $\sum_i R_i = \sum_{kl} R_{kl} = \mathcal{W}_p$, thereby maintaining a strong relation between the input features and the outputted Wasserstein distance.

A second property is its connection to gradient computations for specific choices of $\alpha$ and $\beta$: 
\begin{proposition}
When choosing $\alpha=p$, the relevance scores $R_{kl}$ of pairs of points can be expressed as the gradient computation $R_{kl} = (\partial \mathcal{W}_p / \partial z_{kl}) \cdot z_{kl}$, where we treat $\gamma^\star$ as a constant.
\label{prop:sample-GI}
\end{proposition}
\begin{proposition}
When choosing $\alpha=p$ and $\beta=q$, the relevance scores $R_i$ of the input features can be expressed as the gradient computation $R_{i} = (\partial \mathcal{W}_p/\partial x_{:,i})^\top x_{:,i} + (\partial \mathcal{W}_p / \partial y_{:,i})^\top y_{:,i}$, where $\gamma^\star$ is again treated as constant.
\label{prop:feature-GI}
\end{proposition}
Proofs are given in \supp{C}.
In other words, if we set $\alpha=p$ and $\beta=q$, the LRP rules are expressible as gradient computations. Such a connection is desirable, as the gradient is a local measure of effect and particularly reliable when the function to analyze is smooth. The same gradient identity also justifies our heuristic of capping $\alpha$ and $\beta$ when $p$ and $q$ grow large, and the model becomes increasingly nonlinear. Indeed, the gradient becomes overly local and generally unsuitable for measuring effects in these models of increased nonlinearity.

\subsection{Computational Properties and Implementation}

We briefly discuss the computational requirements of our method. A naive implementation would require maintaining a tensor of size $N \times M \times d$ storing the relevance scores for each explanation (as shown in \cref{fig:wax-attribution-schematic}). This can be bypassed for the case $\alpha=p$ and $\beta=q$ by using the gradient formulations of \cref{prop:sample-GI,prop:feature-GI}. For the more general case where $\alpha \neq p$ or $\beta \neq q$, we can use a ``detach trick''~\cite{samekExplainingDeepNeural2021,aliXAITransformersBetter2022} where some terms of the forward computation are set to constants. Specifically, we replace \cref{eq:fw-1} by:
\begin{center}
\begin{lrpquickbox}
\begin{align}
z_{kl} &= \| x_k - y_l \|_\beta \cdot \left[\frac{\| x_k - y_l \|_q}{\| x_k - y_l \|_\beta}\right]_\mathsf{.detach()}
\label{eq:wax-lrp}
\end{align}
\end{lrpquickbox}
\end{center}
and likewise for \cref{eq:fw-2}. This has the effect of leaving the output of the forward computation unchanged, but making the gradient computations shown in \cref{prop:sample-GI,prop:feature-GI} coincide with those of LRP shown in \cref{eq:wax-attribution-sample,eq:wax-attribution-feature}.
Furthermore, for the case where the transport is sparse (e.g.~classical unregularized transport between distributions with the same number of source and target instances), the structures storing the activations $z_{kl}$ and $R_{kl}$ can be reduced to one-dimensional arrays, further speeding up the computations by a factor $N$.
We provide a Python implementation of \mname{} using the ``detach trick'' in \supp{D} for reference.

Overall, our explanation method is fast and can be applied to large transport models involving many data points and input features without incurring any significant computational overhead.
More generally, unlike other attribution techniques, e.g.~based on feature removal~\cite{strumbeljExplainingPredictionModels2014}, the LRP approach requires only a single evaluation of the prediction function and also avoids the need to address the question of how to remove input features.

\section{Empirical Evaluation}
\label{sec:evaluation}

\begin{table*}[t!]
\caption{SRG scores obtained by each explanation method on various Wasserstein distance models (different values of $p$ and $q$) and datasets. Scores are calculated via mean bootstrap. For each dataset, we also show the number of source and target data points $N$ and $M$, and the number of input features $d$. {\large $*$}~Indicates statistical significance between \mname{} and the next best baseline according to a Wilcoxon test at $p < 0.01$. {\scriptsize \color{gray} \faHourglassStart}~slow runtime. {\scriptsize \color{gray} \faTimesCircle}~equivalent up to a scaling factor to an ablation of \mname{} with $\alpha=1, \beta=2$. Best results are shown in \textbf{bold}.}
\centering
\setlength{\tabcolsep}{7pt}
\begin{tabular}{ccllSSSS}
\toprule
{$\mathcal{W}_p$} & {$\|x-y\|_q$} & {Dataset} &  ($N$\;/\;$M$\;/\;$d$)~~~~~~~~~ & {MeanShift} & {Occlusion$^\text{ \color{gray} \faHourglassStart}$} & {Coupling$^\text{ \color{gray} \faTimesCircle}$} & {\mname{} \textbf{(ours)}} \\
\midrule
\multirow[c]{6}{*}{$p=1$} 
& \multirow[c]{6}{*}{$q=2$} 
& Crime & {(437\;/\;1081\;/\;28)}  & 0.81 & 0.85 & \bfseries 0.90 & \bfseries 0.90 \\
 &  & Mice & {(628\;/\;203\;/\;70)}  & 0.87 & 1.45 & \bfseries 1.46 & \bfseries 1.46 \\
 &  & Musks1 & {(133\;/\;167\;/\;166)} & 0.94 & \bfseries 2.19 & \bfseries 2.19 & \bfseries 2.19 \\
 &  & Robot & {(118\;/\;65\;/\;90)}  & 0.06 & \bfseries 1.48$^{*}$ & 1.48 & 1.48 \\
 &  & Wine & {(3781\;/\;1086\;/\;12)}  & 1.08 & 1.08 & \bfseries 1.09 & \bfseries 1.09 \\
 &  & Wisconsin & {(332\;/\;163\;/\;30)} & \bfseries 1.82 & 1.81 & \bfseries 1.82 & \bfseries 1.82 \\
\midrule
\multirow[c]{6}{*}{$\vdots$} 
& \multirow[c]{6}{*}{$q=1$} 
& Crime &  & 2.49 & 2.54 & 2.42 & \bfseries 2.66${^*}$ \\
 &  & Mice &  & 3.93 & 6.23 & 6.15 & \bfseries 6.24${^*}$ \\
 &  & Musks1 &  & 6.24 & \bfseries 12.85${^*}$ & 10.65 & 12.84 \\
 &  & Robot &  & 3.42 & 9.10 & 3.39 & \bfseries 9.11${^*}$ \\
 &  & Wine &  & \bfseries 2.58${^*}$ & 2.55 & 2.56 & 2.58 \\
 &  & Wisconsin &  & 7.06 & 7.07 & 7.04 & \bfseries 7.08${^*}$ \\
 \cmidrule[0.05pt](l){2-8}
\multirow[c]{6}{*}{$p=2$}  & \multirow[c]{6}{*}{$q=2$} 
& Crime &  & 0.83 & 0.91 & 0.88 & \bfseries 0.96${^*}$ \\
 &  & Mice &  & 0.92 & 1.49 & 1.49 & \bfseries 1.50${^*}$ \\
 &  & Musks1 &  & 1.02 & \bfseries 2.07 & 2.04 & \bfseries 2.07 \\
 &  & Robot &  & 0.64 & \bfseries 1.73 & 1.17 & \bfseries 1.73 \\
 &  & Wine &  & 1.01 & 1.03 & 1.03 & \bfseries 1.04${^*}$ \\
 &  & Wisconsin &  & 1.82 & 1.84 & 1.84 & \bfseries 1.85${^*}$ \\
\cmidrule[0.05pt](l){2-8}
\multirow[c]{6}{*}{$\vdots$} & \multirow[c]{6}{*}{$q=\infty$} 
& Crime &  & 0.35 & 0.54 & 0.25 & \bfseries 0.59${^*}$ \\
 &  & Mice &  & 0.36 & 0.55 & 0.56 & \bfseries 0.59${^*}$ \\
 &  & Musks1 &  & 0.31 & 0.89 & 0.57 & \bfseries 0.97${^*}$ \\
 &  & Robot &  & 0.25 & 0.72 & 0.92 & \bfseries 1.14${^*}$ \\
 &  & Wine &  & 0.47 & 0.50 & 0.45 & \bfseries 0.50${^*}$ \\
 &  & Wisconsin &  & 0.50 & 0.24 & 0.50 & \bfseries 0.54${^*}$ \\
\midrule
\multirow[c]{6}{*}{$p=10$} 
& \multirow[c]{6}{*}{$q=2$} 
& Crime &  & 1.56 & \bfseries 3.57 & 0.91 & \bfseries 3.57 \\
 &  & Mice &  & 0.90 & 1.55 & 1.50 & \bfseries 1.57${^*}$ \\
 &  & Musks1 &  & 0.92 & 2.14 & 1.72 & \bfseries 2.30${^*}$ \\
 &  & Robot &  & 4.81 & 8.29 & -0.68 & \bfseries 8.33${^*}$ \\
 &  & Wine &  & 0.89 & 0.96 & 0.96 & \bfseries 0.98${^*}$ \\
 &  & Wisconsin &  & 1.99 & 2.33 & 2.10 & \bfseries 2.34${^*}$ \\
\bottomrule
\end{tabular}
\label{tab:internal-evaluation}
\end{table*}

Our evaluation procedure consists of two parts. In the first part, \cref{sec:internal-evaluation}, we focus on testing the ability of our \mname{} method to faithfully explain any Wasserstein model, whether it is to gain insight into its predictions or reveal its flaws. In \cref{sec:external-evaluation}, we proceed with a more holistic evaluation, testing the ability of \mname{}---combined with a suitable Wasserstein model---to identify underlying transport phenomena.

\subsection{Evaluating Explanation Faithfulness}
\label{sec:internal-evaluation}

To test the ability of \mname{} to explain Wasserstein distance models faithfully, we use the Symmetric Relevance Gain (SRG) metric~\cite{bluecherDecouplingPixelFlipping2024}\footnote{The SRG is based on the ``pixel-flipping'' evaluation metric~\cite{bachPixelwiseExplanationsNonlinear2015,samekEvaluatingVisualizationWhat2017} but adapted to produce more robust explanation evaluations.}. We denote by $(x_\pi^+, y_\pi^+)$ the dataset of source and target samples obtained by \textit{retaining} only the $\pi$ most relevant features according to the explanation. Likewise, we denote by $(x_\pi^\times, y_\pi^\times)$ the dataset formed by \textit{excluding} the $\pi$ most relevant features according to the explanation. The SRG can then be written as:
\begin{align}
\mathrm{SRG} &= \frac{1}{d+1} \sum_{\pi=0}^{d} \big[\mathcal{W}_p(x_\pi^+, y_\pi^+) - \mathcal{W}_p(x_\pi^\times, y_\pi^\times)\big]
\label{eq:aufc}
\end{align}
where datasets rather than distributions are plugged into $\mathcal{W}_p$ for notational convenience. The SRG intuitively measures the true effect on the Wasserstein distance of adding a feature identified as relevant by the explanation rather than one identified as irrelevant. The higher the SRG, the better the explanation.

\smallskip

We evaluate our method on various specifications of the Wasserstein distance and its underlying Minkowski metric as defined in \cref{eq:wasserstein}, including popular settings such as $\mathcal{W}_1, \mathcal{W}_2$, and the Euclidean distance between points. Corresponding Wasserstein models were built on various datasets of different sizes and dimensionality. To get robust estimates, we repeat each experiment 100 times using bootstrap sampling and show the mean values in \cref{tab:internal-evaluation}.
Further details on the experimental setup, datasets, and their preprocessing are given in Supplementary Notes~E and~F, where we also illustrate the robustness of \mname{}'s attributions for this experiment.

\smallskip

\subsubsection*{Baselines} We contribute a simple set of baselines for benchmarking purposes. The first baseline, \emph{MeanShift}, estimates and compares the mean feature values of the distributions, i.e.~$R_i = (\mathbb{E}[x] - \mathbb{E}[y])^2_i$. This baseline does not consider the model and only looks at the means of the datasets.
The second baseline, \emph{Occlusion}, attributes according to the effect of removing one dimension on the original Wasserstein distance, i.e.~$R_i = \mathcal{W}_p(x, y) - \mathcal{W}_p(x_{:,-i}, y_{:,-i})$, where $x_{:,-i}$ denotes the samples after removing the $i$th dimension (and likewise for $y$). This baseline is highly localized and cannot comprehend the more global effect of removing multiple input features. Moreover, this baseline requires solving the optimal transport function $d$ times (once for every feature removal), which is computationally expensive for large problems. Our last baseline, \textit{Coupling}, looks at the coupling distribution $\gamma^\star$ obtained as a byproduct of calculating the Wasserstein distance. It attributes the distances (assumed to be Euclidean) between the coupled instances to the input features and averages the results over the data. Up to a global scaling factor on the explanation, this baseline is an ablation of \mname{} with its hyperparameters fixed to $\alpha=1$ and $\beta=2$.
In \supp{F}, we theoretically analyze the computational and memory requirements of the evaluated methods and present empirical runtime results.

\smallskip

\subsubsection*{Results}
The results of our evaluation are shown in \cref{tab:internal-evaluation} and complemented in \supp{F}. We observe that our \mname{} method consistently produces the highest SRG scores across Wasserstein models and datasets. This confirms the technical strength of our method, namely its sensitivity to the actual Wasserstein distance model and its ability to attribute the Wasserstein distance comprehensively through its fulfillment of the conservation property (cf.~\cref{tab:baselines-comparison} for a technical comparison between methods).

\begin{table}[b!]
    \setlength{\tabcolsep}{7pt}
    \centering
    \caption{Technical comparison of our proposed \mname{} explanation method against a set of contributed baselines.}
    \begin{tabular}{lcccc}
        \toprule
        Method & Conserving & Model-aware & Efficient \\
        \midrule
        MeanShift & - & - & \checkmark \\
        Occlusion  & - & \checkmark & - \\
        Coupling  & - & (\checkmark) & \checkmark \\
        \mname{} \textbf{(ours)}  & \checkmark & \checkmark & \checkmark \\
        \bottomrule
    \end{tabular}
    \label{tab:baselines-comparison}
\end{table}

Our method outperforms the \textit{MeanShift} and \textit{Coupling} baselines. \textit{MeanShift} suffers from only looking at the data distributions and being insensitive to the actual Wasserstein distance. Likewise, the coupling distribution $\gamma^\star$ used in the \textit{Coupling} baseline inherits the insensitivity of the transport plan to the Wasserstein distance and its parameters (cf.~\supp{A}). The two baselines perform especially weakly compared to \mname{} when explaining Wasserstein models with high $p$ and $q$ values, as they fail to account for the high sensitivity of these Wasserstein models to specific outlier instances or features. In contrast, \mname{} can finely adapt the exposure of the explanation to outliers through a judicious choice of parameters $\alpha$ and $\beta$.

\mname{} also outperforms the \textit{Occlusion} baseline, despite the latter being much more computationally expensive. By only simulating one feature removal step, \textit{Occlusion} cannot account for the effect on the Wasserstein distance of removing multiple features. The locality of the \textit{Occlusion} is particularly problematic for models with sharp non-linearities near joint feature activations (such as those with large $q$), whereas \mname{} is better equipped to handle these highly nonlinear cases through the capping heuristic for the parameter $\beta$ and the smoothing it induces on the explanations.
This advantage could, e.g., aid in analyzing persistence diagrams with the Wasserstein distance, where the `bottleneck distance' ($p,q=\infty$) is used.

Overall, the results correlate with the fulfillment of the technical desiderata studied in \cref{tab:baselines-comparison} and the empirical performance reported in \cref{tab:internal-evaluation}. Our method, which fulfills all the considered desiderata, also performs the best empirically.

\subsection{Characterization of Transport Phenomena}
\label{sec:external-evaluation}

\begin{table*}[b!]
\centering
\setlength{\tabcolsep}{8pt}
\caption{
Mean cosine similarities between the ground-truth relevance scores from \cref{eq:relevance-ground-truth}, and relevance scores computed by each method. For each dataset, we also show the median number of source and target data points $\tilde{N}$ in each subset $\mathcal{D}_\mathcal{S}, \mathcal{D}_\mathcal{T}$, and the number of input features $d$. Best results are shown in \textbf{bold}.
}
\begin{tabular}{llSSSSSS}
\toprule
 & & & & & & \multicolumn{2}{c}{\mname{} \textbf{(ours)}} \\
\cmidrule(lr){7-8}
{Dataset} & ($\tilde{N}$\;/\;$d$) & {Uniform} & {MeanShift} & {Logistic [$\nabla^2$]} & {Logistic [GI]} & {$\gamma^\star$} & {$\gamma^\star_\mathrm{reg}$}\\
\midrule
Air Quality (every 1h) & ($363$\;/\;$9$) & 0.83 & 0.82 & 0.65 & 0.47 & \bfseries 0.85 & 0.80 \\
Air Quality (every 2h) & ($354$\;/\;$9$) & 0.84 & 0.87 & 0.68 & 0.40 & \bfseries 0.91 & 0.83 \\
Air Quality (every 3h) & ($353$\;/\;$9$) & 0.85 & 0.88 & 0.69 & 0.40 & \bfseries 0.93 & 0.87 \\
Air Quality (every 4h) & ($349$\;/\;$9$) & 0.86 & 0.87 & 0.67 & 0.40 & \bfseries 0.94 & 0.89 \\
Air Quality (every 5h) & ($351$\;/\;$9$) & 0.87 & 0.92 & 0.67 & 0.43 & \bfseries 0.95 & 0.90 \\
Air Quality (every 6h) & ($352$\;/\;$9$) & 0.88 & 0.91 & 0.67 & 0.42 & \bfseries 0.95 & 0.91 \\
\midrule
Electricity (every 1h) & ($1213$\;/\;$7$) & 0.84 & 0.61 & 0.47 & 0.43 & 0.83 & \bfseries 0.88 \\
Electricity (every 2h) & ($1201$\;/\;$7$) & 0.90 & 0.68 & 0.54 & 0.42 & 0.86 & \bfseries 0.94 \\
Electricity (every 3h) & ($1192$\;/\;$7$) & 0.93 & 0.71 & 0.55 & 0.35 & 0.86 & \bfseries 0.96 \\
Electricity (every 4h) & ($1191$\;/\;$7$) & 0.94 & 0.74 & 0.60 & 0.42 & 0.87 & \bfseries 0.97 \\
Electricity (every 5h) & ($1221$\;/\;$7$) & 0.94 & 0.75 & 0.60 & 0.37 & 0.88 & \bfseries 0.98 \\
Electricity (every 6h) & ($1234$\;/\;$7$) & 0.94 & 0.77 & 0.61 & 0.37 & 0.88 & \bfseries 0.98 \\
\midrule
Appliances (every 1h) & ($124$\;/\;$25$) & 0.38 & 0.28 & 0.06 & 0.10 & \bfseries 0.51 & 0.41 \\
Appliances (every 2h) & ($121$\;/\;$25$) & 0.47 & 0.35 & 0.10 & 0.14 & \bfseries 0.62 & 0.50 \\
Appliances (every 3h) & ($120$\;/\;$25$) & 0.53 & 0.40 & 0.15 & 0.19 & \bfseries 0.69 & 0.56 \\
Appliances (every 4h) & ($119$\;/\;$25$) & 0.58 & 0.43 & 0.20 & 0.25 & \bfseries 0.74 & 0.62 \\
Appliances (every 5h) & ($119$\;/\;$25$) & 0.62 & 0.48 & 0.23 & 0.29 & \bfseries 0.78 & 0.66 \\
Appliances (every 6h) & ($119$\;/\;$25$) & 0.65 & 0.54 & 0.26 & 0.34 & \bfseries 0.81 & 0.69 \\
\midrule
PLISM (Ciga-ResNet) & ($5000$\;/\;$512$) & 0.70 & 0.90 & 0.59 & 0.64 & \bfseries 1.00 & 0.92 \\
PLISM (Phikon) & ($5000$\;/\;$768$) & 0.43 & 0.96 & 0.90 & 0.94 & \bfseries 1.00 & 0.92 \\
PLISM (UNI) & ($5000$\;/\;$1024$) & 0.83 & 0.77 & 0.73 & 0.75 & \bfseries 0.99 & 0.93 \\
\bottomrule
\end{tabular}
\label{tab:external-evaluation}
\end{table*}

We now examine the ability of \mname{}, when paired with a suitable Wasserstein distance model, to accurately characterize transport phenomena, specifically the dimensions along which transport occurs. We first artificially generate transport phenomena with ground-truth from three multivariate time series datasets, `Air Quality', `Electricity', and `Appliances', as described in \supp{E}. Denoting the multivariate time series $x_1, x_2, \dots$ with periodicity $T$, we generate the source and target distributions
\begin{align*}
\mathcal{D}_\mathcal{S} &= \big\{(x_{t+kT})\big\}_{k \in \mathbb{Z}}\\
\mathcal{D}_\mathcal{T} &= \big\{(x_{t+\Delta t+kT})\big\}_{k \in \mathbb{Z}}
\end{align*}
where $t$ is the source time (e.g.~$7$\,am in the morning), $t + \Delta t$ is the target time (e.g.~$9$\,am in the morning, i.e.~two hours later), and $T$ is the periodicity of the time series ($24$ hours in our experiments). We only retain coupled points from the same day, i.e.~$0 \leq t < t + \Delta t \leq 23$. A ground-truth estimate of the transport along a specific dimension $i$ can then be obtained by computing:
\begin{align}
R_i^\text{(true)} &= \mathbb{E}_k\big[ (x_{t+\Delta t + kT,i} - x_{t+kT,i})^2 \big]
\label{eq:relevance-ground-truth}
\end{align}
We set as an objective for \mname{} to reconstruct these ground-truth relevance scores, specifically, producing a vector of relevance scores with maximum cosine similarity to the ground-truth, while having only access to samples from the sets $\widetilde{\mathcal{D}}_\mathcal{S} = \{(x_{t+kT})\}_{k \in \mathcal{K}_1}$ and $
\widetilde{\mathcal{D}}_\mathcal{T} = \{(x_{t+\Delta t+kT})\}_{k \in \mathcal{K}_2}$.
Details of the experiment and processing of these datasets are given in Supplementary Notes~E and~G.

Apart from the time series data, we also use the PLISM~\cite{filiotDistillingFoundationModels2025} histopathology dataset.
It consists of a fixed set of patches from various tissue types, each recorded multiple times with different stainings and scanners.
We use this natural coupling and evaluate two different staining variations of the same patches and embed these into the latent spaces of the histopathology foundation models `Ciga-ResNet'~\cite{cigaSelfSupervisedContrastive2022}, `Phikon'~\cite{filiotScalingSelfSupervisedLearning2024}, and `UNI'~\cite{chenGeneralpurposeFoundationModel2024}.
These models differ in recency, size, and performance.
The resulting embeddings are each normalized to unit norm.

We consider two variants of \mname{} using $p=2$ and $q=2$: one where the coupling $\gamma^\star$ is the exact solution of \cref{eq:wasserstein}, and one where the coupling is set uniform, i.e.~fixed to $\gamma^\star_{\mathrm{reg}} = \bone_{N \times N} / N^2$. This variant can be seen as a limit case of the Sinkhorn~\cite{cuturiSinkhornDistancesLightspeed2013} distance (a regularized Wasserstein distance) with maximum entropic smoothing and additionally as a more informed \textit{MeanShift} that includes the variance of the data.

\smallskip

\subsubsection*{Baselines}
We consider three baselines. The simplest is \textit{Uniform}, which assigns relevance uniformly and constantly as $R_i = 1$.
This baseline, therefore, represents an equal impact of all features.
We also include \textit{MeanShift}, as already defined in \cref{sec:internal-evaluation}, and a \textit{Logistic} baseline, which derives explanations from a linear classifier $f(x) = w^\top x + b$, trained using logistic regression. Attributions are computed via sensitivity and GI: $R^{\nabla^2}_i = (\partial f / \partial x_i)^2 = w_i^2$ and $R^{\mathsf{GI}}_i = w_i (\E[x] - \E[y])_i$ (cf.~\supp{H}). 
The source and target domains are used as a binary classification problem in this case. A particularly interesting aspect about the \textit{MeanShift} and \textit{Logistic} baselines is that they can be seen as specific cases of a recent proposal based on attributing the symmetrized Kullback-Leibler divergence (also known as `Jeffrey's divergence') $D_\mathsf{KL}(\mu, \nu) = D_\mathsf{KL}(\mu\parallel\nu) + D_\mathsf{KL}(\nu\parallel\mu)$ between the two distributions (cf.~\supp{H}). A related KLD-based analysis can be found in~\cite{budhathokiWhyDidDistribution2021} in the context of causal graphs.

\smallskip

\subsubsection*{Results}
In \cref{tab:external-evaluation}, we see that \mname{} systematically ranks first in characterizing the ground-truth transport phenomena. In particular, our method outperforms \textit{MeanShift}, which represents translations in input space, but no increases/decreases in variance or more complex distortions. These more complex forms of distributional change appear to dominate across the transport phenomena, and this effect is consistent across various considered delays $\Delta t$. The \textit{Logistic} baseline performs significantly worse on this benchmark, regardless of whether sensitivity analysis ($\nabla^2$) or Gradient\,$\times$\,Input (GI) is used for the final explanation step. This can be explained by a mismatch between the directions of actual transport and the directions that discriminate between sources and targets. This mismatch can be shown to occur as soon as the source and target distributions are not isotropic.
A related phenomenon is discussed in~\cite{haufeInterpretationWeightVectors2014}.
More generally, classifier-based approaches struggle when the source and target distributions are highly overlapping, which is not a problem for the Wasserstein distance model.

\section{Use Case 1: Identifying Robust Features for Aligning Domains}
\label{sec:use-case-relevant-feature-discovery}
To demonstrate the tangible benefits \mname{} can deliver in a real-world application, we explore a domain adaptation scenario, building on the increasingly dominant paradigm of constructing classifiers on top of representations provided by `foundation models'.  While this approach has led to significant performance gains across a wide range of ML tasks, recent work (e.g.~\cite{kauffmannExplainableAIReveals2025, komenRobustFoundationModels2025}) has found that it is less robust than previously thought.
In particular, the self-supervised nature of foundation models makes them prone to picking up domain-specific signals, and relying on them can severely degrade domain adaptation capabilities and downstream prediction performance.
Such behavior can have serious consequences in sensitive areas such as medicine, where hospital-specific biases (i.e.~`batch effects') may be picked up in favor of the actual biological signal~\cite{komenRobustFoundationModels2025}.

In this use case, we illustrate \mname{}'s ability to address domain adaptation by identifying domain-specific features. Pruning these from the input space shifts the classifier's reliance to domain-invariant signals, thereby improving its robustness.
To this end, we propose the following scoring rule based on \mname{}'s feature attributions:
\begin{align}
    R^{\mathsf{DA}}_i &= R_i \big/ \sigma_i^{\lambda}
    \label{eq:wax-da}
\end{align}
where $R_i$ is the standard \mname{} attribution (\cref{eq:wax-attribution-feature}), and $\sigma_i$ represents the pooled standard deviation of feature $i$ across both domains. Pruning features in decreasing order of \cref{eq:wax-da} removes input directions that strongly contribute to the transport cost between domains, while keeping features that maintain the variance within domains. The hyperparameter $\lambda \in [0, \infty)$ controls the trade-off between domain-discriminative versus high-variance directions.
Intuitively, $\lambda = 0$ recovers the base \mname{} attribution, focusing solely on the direction of maximum domain discrepancy. Increasing $\lambda$ aligns with the principles of `Relevant Dimensionality Estimation'~\cite{braunRelevantDimensionsKernel2008}, as $\lambda$ weighs domain-discriminative signals against high-variance directions, thereby limiting the impact of low-varying, artifact-prone features. In the limit as $\lambda \to \infty$, this method relates to PCA, pruning features purely based on their variance.
An ablation study analyzing the effect of $\lambda$ is provided in \supp{I}.

\begin{figure*}[!t]
    \centering
    \includegraphics[width=\linewidth]{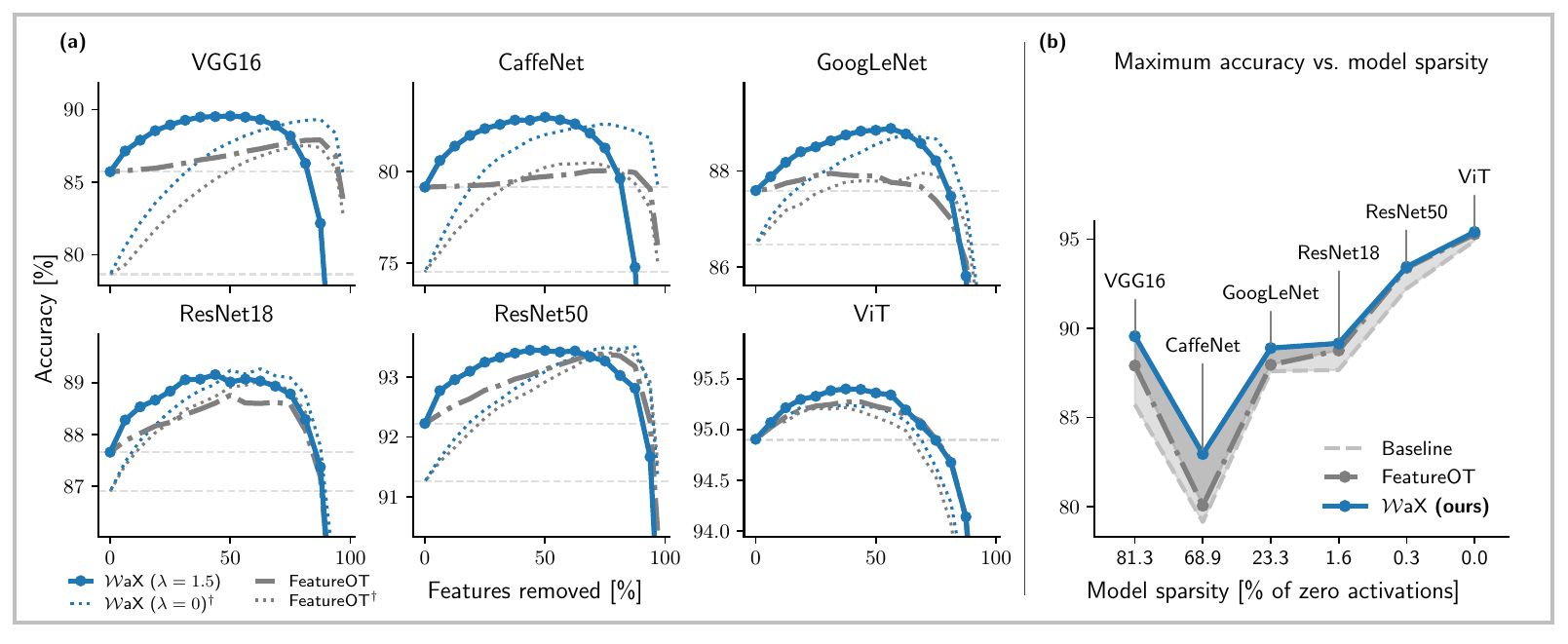}
    \caption{
    Averaged 1-NN accuracies after aligning domain pairs on feature subsets with \mname{} and \textit{FeatureOT}~\cite{gautheronFeatureSelectionUnsupervised2018}. (a)~Accuracies after removing the most relevant $\times\%$ of features according to the methods. (b)~Maximum accuracy for each method and feature space in comparison to the sparsity of the feature spaces (\% of zero activations). Both \mname{} and \textit{FeatureOT} show consistent improvements over the baseline performance (using all available features). Moreover, \mname{} handles sparsity more efficiently and consistently yields the highest accuracies for all models. {$\dagger$:}~Evaluation on separately scaled domains.}
    \label{fig:use-case-domain-adaptation}
\end{figure*}

Our approach aligns broadly with OT-based domain adaptation methods (cf.~\cite{courtyOptimalTransportDomain2017}), and more specifically, with Gautheron {\em et al.}~\cite{gautheronFeatureSelectionUnsupervised2018}, who propose an OT-based feature selection strategy that identifies shared features across two domains.
They showed that domain alignment benefits from using a small subset of robust features, thereby improving domain adaptation capabilities.
Pruning the feature space can mitigate the adverse effects of uninformative features, biases, and spurious (domain-specific) correlations, leading to more robust, task-specific representations.
In the following, we show that \mname{} can be applied to align domains and handle sparse feature representations.

We adopt the experimental setup proposed by Gautheron {\em et al.}~\cite{gautheronFeatureSelectionUnsupervised2018} and apply it to the `Office-Caltech10' domain adaptation benchmark~\cite{gongGeodesicFlowKernel2012}.
This benchmark consists of images from four domains (\textit{Amazon}, \textit{Caltech10}, \textit{DSLR}, and \textit{Webcam}) that share the same ten classes.
We compare \mname{} against \textit{FeatureOT}~\cite{gautheronFeatureSelectionUnsupervised2018} on six ImageNet-pretrained neural network feature representations with varying levels of sparsity (cf.~\cref{fig:use-case-domain-adaptation}).
Two evaluation settings are considered: `unscaled' (raw data) and `scaled' (each domain separately standardized to zero mean and unit variance).
While `unscaled' tests the most general setting, `scaled' provides statistically pre-aligned representations.
Alignment quality is measured using 1-Nearest-Neighbor accuracy on the target data after pruning features based on their attribution scores.
We report averages over 20 repetitions for all 12 domain pairs (more details in~\cite[Sec.~5.1]{gautheronFeatureSelectionUnsupervised2018}).
Note that no subsequent domain adaptation algorithm is applied. Our goal is to directly isolate and evaluate \mname{}'s ability to identify robust, domain-invariant features.
Thus, \mname{} serves as an interpretable preprocessing step to improve efficiency and robustness, rather than a direct competitor to full domain adaptation pipelines.

\Cref{fig:use-case-domain-adaptation}a shows the feature-removal curves across all domain pairs for both `unscaled' (markers) and `scaled' (no markers) feature spaces.
\mname{} successfully identifies domain-specific features, as indicated by the immediate accuracy gain over the baseline accuracy (dashed horizontal lines) after their removal.
Notably, \mname{} demonstrates robustness across both settings, achieving the highest performance on the unscaled data without requiring statistical preprocessing.
\textit{FeatureOT}, on the other hand, is more competitive in the scaled setting.
\Cref{fig:use-case-domain-adaptation}b further emphasizes the performance gap between \textit{FeatureOT} and \mname{}, particularly in terms of sparsity of the representations and in the more general unscaled modality.
Specifically, we plot the maximum accuracy achieved by each method over all pruning steps. 
We observe a clear trend where the accuracy gap between \mname{} and \emph{FeatureOT} widens with increased sparsity, underscoring \mname{}'s efficacy in handling high-dimensional, sparse representations.

Overall, we have demonstrated the practical utility of \mname{} for identifying and pruning non-robust, domain-specific features. 
By isolating domain-invariant components, our method enables more robust downstream classifier performance, with the added advantage of being easy to deploy and interpretable.

\section{Subspace-Based Explanations}
\label{sec:concept-based-xot}

We now introduce ``\wdsa{}'': an extension of \mname{} aiming at producing a new form of explanation, namely, an attribution of the Wasserstein distance on subspaces of the input domain. These subspaces will be set to be interpreted as abstract concepts describing sub-shifts. The proposed method restricts itself to the specific choice of parameters $p=2$ and $q=2$, corresponding to the common $\mathcal{W}_2$ distance with underlying Euclidean metric for pairwise comparisons. The starting point of \wdsa{} is to define an orthogonal matrix $\boldsymbol{U}$ of size $d \times d$ that we decompose column-wise into blocks: 
\begin{align}
\boldsymbol{U} &= (U_1\,|\dots|\,U_C\,|\,U_\perp)
\label{eq:orthogonal}
\end{align}
where the first $C$ blocks define $C$ subspaces of the input domain, each associated with a concept, and the last block $U_\perp$ is the orthogonal complement. Note that concepts can be associated with subspaces of potentially different sizes. Under such a decomposition of the input space, we observe that the Wasserstein distance can be expressed as a three-layer neural network:
\begin{center}
\begin{fwbox}
\begin{subequations}
\label{eq:wd-subspace}
\begin{align}
z_{kl}^c &= \|U_c^\top (x_k - y_l)\|_2 \label{eq:fwx-1}\\
S_c &= \Big(\sum_{kl} \gamma_{kl}^\star \big(z_{kl}^c\big)^2\Big)^{1/2} \label{eq:fwx-2}\\
\mathcal{W}_2 &= \Big(\sum_c S_c^2\Big)^{1/2} \label{eq:fwx-3}
\end{align}
\end{subequations}
\end{fwbox}
\end{center}
where the second layer outputs concept-specific scores, and the last layer performs a pooling operation over these concepts.
We stress that the output of \cref{eq:fwx-3} is equivalent to the output of \cref{eq:fw-2} with $p=2$ and $q=2$, and therefore we are still considering the same original Wasserstein distance.

\medskip

This more sophisticated restructuring of the Wasserstein distance allows it to be attributed to the concepts defined by the newly defined subspaces. In particular, we propose an alternate set of LRP rules, which proceeds first by redistribution onto those concepts, \cref{eq:subspace-attribution-concepts}, then the data points, \cref{eq:subspace-attribution-sample}, and finally the input dimensions \cref{eq:subspace-attribution-features}:
\begin{center}
\begin{lrpbox}
\begin{subequations}
\label{eq:wax-subspace}
\begin{align}
R_c &= \frac{S_c^2}{\sum_{c} S_c^2} \mathcal{W}_2 \label{eq:subspace-attribution-concepts}\\
R_{kl}^c &= \frac{\gamma_{kl}^\star (z^c_{kl})^2}{\sum_{kl} \gamma_{kl}^\star (z^c_{kl})^2} R_c \label{eq:subspace-attribution-sample}\\
R_i^c &= \sum_{kl} \frac{(x_{ki} - y_{li}) \cdot (\hat{x}_{ki} - \hat{y}_{li})}{\sum_{i} (x_{ki} - y_{li}) \cdot (\hat{x}_{ki} - \hat{y}_{li})} R^c_{kl}
\label{eq:subspace-attribution-features}
\end{align}
\end{subequations}
\end{lrpbox}
\end{center}
where $(\hat{x}_k - \hat{y}_l) = U_c U_c^\top (x_k - y_l)$.
As with standard \mname{}, the LRP conservation property still holds.

What we have left unanswered so far is how to generate an orthogonal matrix $\boldsymbol{U}$ whose subspaces carry a specific meaning (e.g.~a concept or sub-shift). In the following, we extend the approach of~\cite{chormaiDisentangledExplanationsNeural2024}, which addresses the question of finding those subspaces in the context of neural networks. Inspired by subspace analysis methods such as Independent Component Analysis (ICA), we proceed by associating to each concept $c$, a tailedness statistic:
\begin{align}
Q_c &= \Big(\sum_{kl} \gamma_{kl}^\star \big(z_{kl}^c\big)^r\Big)^{1/r}
\label{eq:subspace-tailedness}
\end{align}
where $r \geq 2$ is a hyperparameter. Setting $r$ large corresponds to letting the largest distances between coupled points dominate the statistic. Having set the hyperparameter $r$, we then proceed by searching for the subspaces that maximize the tailedness statistics of each subspace:
\begin{align}
\max_{\boldsymbol{U}} \big\{Q_1 + \dots + Q_C \big\}
\label{eq:subspace-objective}
\end{align}
subject to $\boldsymbol{U}$'s orthogonality. Optimization can be achieved by alternating gradient ascent and orthogonalization steps. Note that for the special case where $r=2$ and $C = 1$, the solution can be found more directly in the eigenvectors of the matrix $\sum_{kl} \gamma_{kl}^\star (x_k-y_l) (x_k-y_l)^\top$ (see \supp{J}).

It is important to stress that, unlike classical subspace analyses such as PCA and ICA, our proposed \wdsa{} does not simply describe the data. Instead, it focuses on directions that are \textit{relevant} to the transport phenomena. Conversely, directions of large variance are ignored by our method if they are present both in the source and target data. 
Additionally, the parameter $r$ allows us to control structural properties of these directions. While the global aggregation in \cref{eq:fwx-3} requires the $\ell_2$-norm to preserve the Wasserstein distance under rotations, setting $r>2$ rewards localization on a few dominant instance pairs, thereby disentangling distinct sub-shifts. 
Subspace analyses that focus on the target task are a recent direction in Explainable AI (e.g.~\cite{chormaiDisentangledExplanationsNeural2024}), and we hereby provide an instantiation of it to the case of Wasserstein distances.

\section{Use Case 2: Exploring an Aging Phenomenon}
\label{sec:use-case-abalone-age}

In the following, we will demonstrate how our \mname{} framework, in particular \wdsa{}, can shed light on a real-world transport phenomenon through only accessing data at the distribution level. Specifically, we study the scenario of a heterogeneous cohort observed twice (once at time $t$, and once at time $t+\tau$), and want to characterize how it evolves in this time interval. We simulate such a scenario using the abalone~\cite{nashAbalone1995} dataset. The number of rings on an abalone roughly represents its age in years ($\text{age} \approx \text{rings}+1.5$)~\cite{nashAbalone1995}. We use this relation to generate two subsets of data spanning the ranges $6$ to $14$ rings and $7$ to $15$ rings, and sampled in a way that simulates the same cohort being observed twice at a one-year interval (see \supp{E} for preprocessing details). We refer to these sets as source and target.
Note that, since for this concrete example there is no known ground truth available, unlike in \cref{sec:external-evaluation}, this use case is not intended to serve as an evaluation of retrieving the \textit{true} transport phenomenon, but rather to illustrate the conceptual aspects of \wdsa{} in comparison to alternative approaches.

\begin{figure*}[!t]
    \centering
    \includegraphics[width=\linewidth]{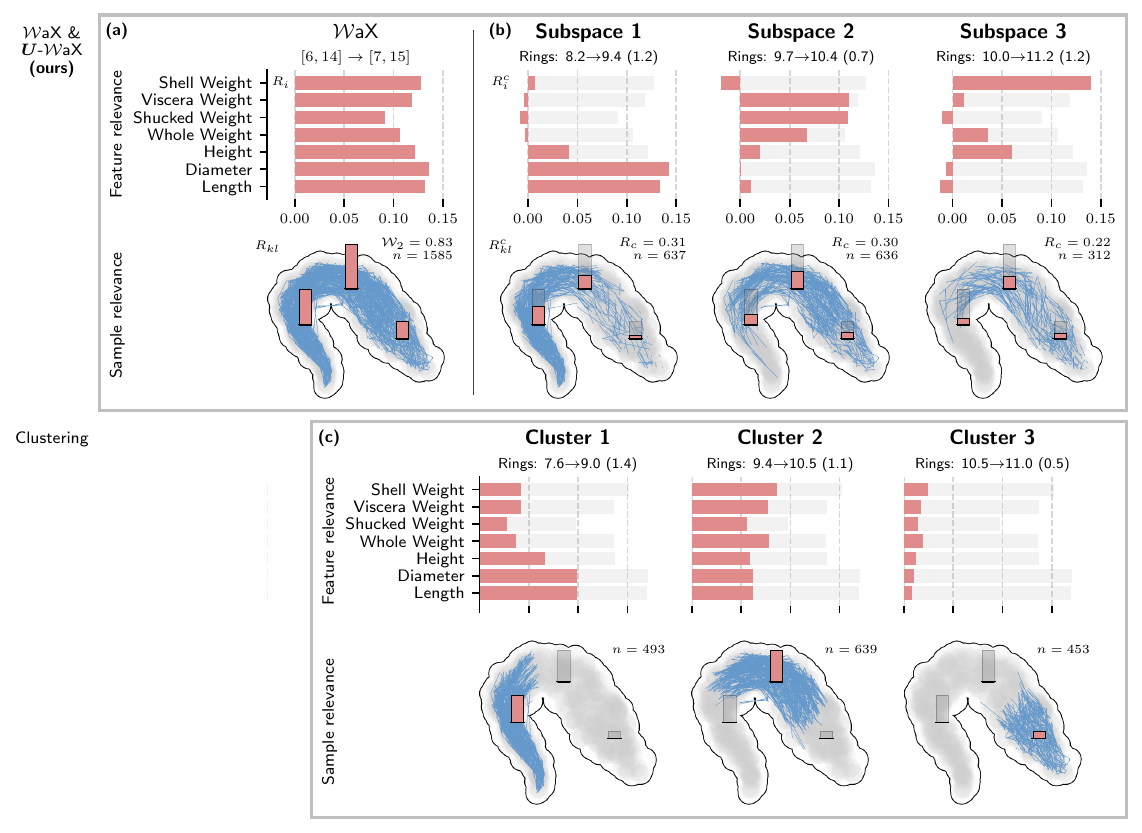}
    \caption{Insight into an abalone aging process through the \mname{} and \wdsa{} methods. The Wasserstein distance models the shift in a simulated abalone cohort measured twice at approximately a one-year interval. (a)~Instance-wise ($R_{kl}$) and feature-wise attributions ($R_i$) calculated by \mname{}. (b)~Attribution on subspaces (i.e.~$R^c_{kl}$ and $R^c_i$) extracted by \wdsa{} ($r=4$). (c)~Similar analysis where \wdsa{} is substituted by a simpler clustering baseline~\cite{kulinskiExplainingDistributionShifts2023} to generate `subspaces'. Instance-wise attributions are summarized in three data subgroups based on the cluster assignments of (c) and overlaid on the t-SNE plot of the data.
    Transport arrows in each t-SNE show the coupled sample pairs that are strongest in the subspace, or assigned to the cluster.
    The bar sizes indicate the contribution of each data component to the Wasserstein distance.
    }
    \label{fig:use-case-abalone-age}
\end{figure*}

To comprehend the modeled aging process, we build a Wasserstein distance model between source and target, with $p=2$ and $q=2$. The modeled Wasserstein distance subsumes the change in feature space resulting from the cohort aging by one year. We proceed with the application of \mname{}, which allows us to uncover features that contribute most to the distributional change modeled by the Wasserstein distance. Results are shown in \cref{fig:use-case-abalone-age}a, where vertical and horizontal bars denote instance-wise and feature-wise contributions to the Wasserstein distance, as computed by \cref{eq:wax}. The results highlight that all instances and features are relevant to aging.

While the analysis above provides fundamental insight, we note that the simulated abalone cohort is highly heterogeneous, spanning a wide range of ages and containing different abalone subtypes. Aging in this cohort is likely a multifaceted process with distinct subgroups of abalones aging differently. To investigate this hypothesis, we apply our proposed \wdsa{} analysis and configure it to decompose the Wasserstein distance into three subspaces. The result of this more detailed analysis is presented in \cref{fig:use-case-abalone-age}b, where vertical and horizontal bars again denote the instance-wise and feature-wise contributions, this time filtered by each subspace, i.e.~computed using \cref{eq:subspace-attribution-features,eq:subspace-attribution-sample}, respectively.

Our analysis confirms the multifaceted nature of the modeled aging process, where different subgroups of instances are associated with different features. For example, contrasting subspace $S_1$ with subspaces $S_2$ and $S_3$ reveals that the height, diameter, and length features tend to evolve and contribute distinctly to weight-related features in the overall abalone aging process. This partitioning of features discovered by our analysis coincides with a well-known physical law: weight tends to increase nonlinearly with length or diameter, specifically following some power of these quantities. Concretely, it follows from this law that weight variations associated with a given increase in diameter are significantly bigger in large abalones than in small ones. Lastly, the differences between $S_2$ and $S_3$ reveal some heterogeneity in how abalones acquire weight, with weights of different parts of the abalone exhibiting relatively independent growth. This latter observation could be linked to the manifestation of distinct subtypes among larger abalones, each of which may follow distinct temporal evolutions.

To highlight the unique disentanglement capabilities of \wdsa{}, we show for comparison a similar analysis where it is substituted by the clustering approach proposed in~\cite{kulinskiExplainingDistributionShifts2023}. Their algorithm first computes the optimal transport map between source and target distributions. It then performs $k$-means clustering on the column-wise concatenated source $x$ and transported $y$ vectors $(x,y)$, resulting in a set of prototypes $(m_c^x,m_c^y)_{c=1}^C$, where the differences $\delta_c = m_c^y-m_c^x$ model the mean shift in each cluster. For comparability with \wdsa{}'s attribution, we model every transported instance pair $(x,y)$ by the shift of its associated cluster, allowing us to formulate the problem of explanation as that of attributing $\sum_c \sum_{k \in c} \|\delta_{c}\|^2$. This quantity naturally decomposes into feature-wise relevances $R^c_i = \sum_{k \in c} (\delta_c)^2_i$ and instance-wise relevances $R^c_{k} = 1_{k \in c}\|\delta_{c}\|^2$. Results of this baseline attribution technique are shown in \cref{fig:use-case-abalone-age}c.

Compared to \wdsa{}, we observe that the clustering baseline produces maximal disentanglement in terms of \textit{instances} (a consequence of the cluster-based formulation), but low feature-wise disentanglement, not clearly distinguishing between weight-related and size-related features.

\medskip

Overall, this use case demonstrated the abilities of \wdsa{} to unravel the shift behind a simulated aging process by only having access to distributions at two discrete times. Through its disentanglement capabilities \wdsa{} supplements \mname{} in producing more structured and insightful explanations of heterogeneous transport phenomena.

\section{Use Case 3: Investigating Dataset Differences}
\label{sec:use-case-dataset-differences}

\begin{figure*}[tb!]
    \centering
    \includegraphics[width=.95\linewidth]{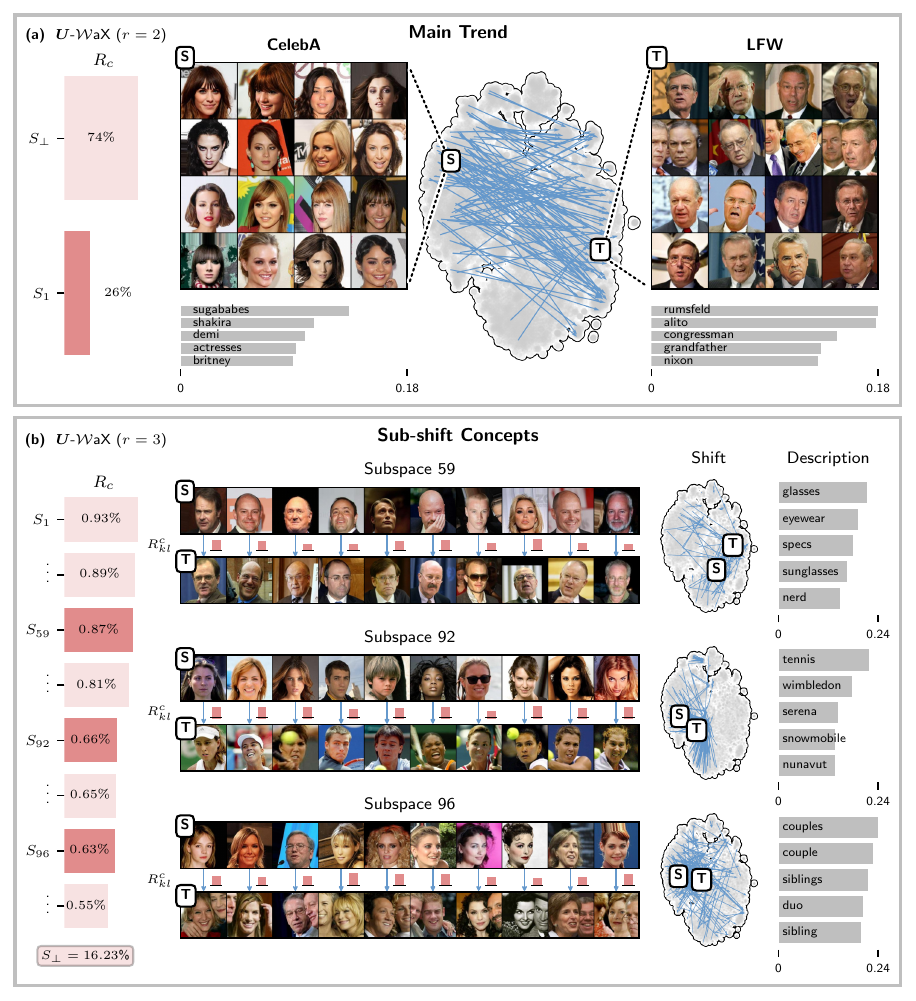}
    \caption{
        Shift between the CelebA and LFW datasets as modeled by the Wasserstein distance ($p,q=2$) and characterized by \wdsa{}. We show \wdsa{} with $r=2$ in~(a) and with $r=3$ in~(b). While $r=2$ provides a high-level/global view of the shift by modeling the main transport trend, $r=3$ reveals diverse local sub-shifts.
        For ease of visualization, the subspace relevances $R_c$ from \cref{eq:subspace-attribution-concepts} are provided as an average over the intermediate range when labeled with vertical dots, and $S_\perp$ indicates the unexplained residual relevance.
        The bars between the coupled samples indicate the sample relevance $R^c_{kl}$ from \cref{eq:subspace-attribution-sample} within each subspace. For each subspace, we also show the t-SNE plots visualizing the localization and spread of the shifts, and the words with the highest absolute cosine similarity.
    }
    \label{fig:celeba-lfw-subshifts}
\end{figure*}

Selecting data sources is an essential aspect of learning an ML model. Thus, it is necessary to systematically evaluate in which way the various data sources differ (e.g.~\cite{alvarez-melisGeometricDatasetDistances2020,babbarWhatDifferentThese2025}) to detect, for example, whether one dataset under- or overrepresents specific subgroups. In this section, we show how \wdsa{} can be used to explore qualitatively the difference between two popular face datasets, namely `CelebA'~\cite{liuDeepLearningFace2015} and `Labeled Faces in the Wild' (LFW)~\cite{huangLabeledFacesWild2007}. To focus on the semantically significant difference between the datasets, we apply rescaling to images so that the faces are centered and of the same size. We then randomly draw \num{10000} examples from each dataset and feed them into a pretrained CLIP~\cite{radfordLearningTransferableVisual2021} model to generate latent features. Not only does the CLIP model produce a high-level abstract representation of images from which semantically interesting transport phenomena can be extracted, but the representation jointly expresses visual and textual features, enabling the characterization of the transport phenomena in terms of words. Our complete feature extraction pipeline involves center-cropping the LFW images to reduce background content (as CelebA images are already cropped to facial regions), applying the CLIP model's prescribed preprocessing transforms, and normalizing each resulting embedding to unit norm.

We first apply our \wdsa{} method and search for a one-dimensional subspace that captures most variation in the modeled transport to get an impression of the main transport trend. Specifically, we apply our method with parameters $C=1$ and $r=2$ in \cref{eq:subspace-tailedness,eq:subspace-objective}, and set the subspace dimensionality to $1$. The resulting subspace captures about 26\% of the Wasserstein distance. Results of the analysis are shown in \cref{fig:celeba-lfw-subshifts}a, where we show the $16$ coupled instances that maximally contribute to the Wasserstein distance as per \cref{eq:subspace-attribution-sample}. This observed transport trend is a clear indicator of differences in the representation of various demographics across the two datasets, in particular, a lower representation of (young) females in LFW compared to CelebA. To get further insight into this transport trend, we leverage the joint representation space of CLIP and search for word vectors that maximally align with this subspace, specifically, with maximum cosine similarity. We search among the \num{20000} most frequent English words\footnote{\url{https://github.com/first20hours/google-10000-english}} (as also used by~\cite{oikarinenCLIPdissectAutomaticDescription2023}). We observe that the most aligned words are on the one hand (female) celebrities, which is supported by multiple female names and words like `{actresses}'; on the other hand, politics-focused words, such as male politician names, `{congressman}', and `{grandfather}'.

\medskip

Aiming at obtaining finer and more structured insights into dataset differences, we apply the same method but choose $C=100$ (with subspaces of one dimension) and $r=3$ to extract a broader range of trends that are more specific to particular subsets of the data. These $100$ subspaces capture $84\%$ of the total contributions to the Wasserstein distance. Three of the learned subspaces are shown in \cref{fig:celeba-lfw-subshifts}b.
Here, we show the coupled samples that most distinctly contribute to the Wasserstein distance through the particular subspace, specifically, having the maximum relevance ratio between the given subspace and the next most relevant subspace.

The first shown subspace, $S_{59}$, in \cref{fig:celeba-lfw-subshifts}b appears to model a shift in wearing `{eyewear}', which is characterized by the highest cosine similarities to related words and the visual characteristics of the strongest sample pairs (i.e.~the lack of `{glasses}' in the source set). This shift is relatively independent from the main female-to-male shift highlighted in \cref{fig:celeba-lfw-subshifts}a, as the insertion of glasses does not come with substantial variation in the characteristics of the faces on which these glasses are inserted.
Subspace $S_{92}$ highlights another subtle difference between the CelebA and LFW datasets, where neutral faces, mostly of women, transport to similar-looking faces adorned with tennis-related items, such as rackets, balls, and outfits. The nature of this sub-shift is confirmed by words such as `{tennis}', `{wimbledon}', and `{serena}' (a famous tennis player), exhibiting the highest cosine similarity to the subspace. As with the previous sub-shift, this sportswear sub-shift appears largely independent of the dominant shift, as coupled instances do not reveal additional differences.
Looking now at the most strongly contributing examples in $S_{96}$, we see that LFW differentiates itself from CelebA by the presence of images containing more than one person. This observation is supported by the strong alignment of the subspace with corresponding text attributes like `{couples}', indicating that the subspace aligns with the visual information provided by the highly relevant samples. If a classifier is trained on datasets with only a single visible face, such samples can be ambiguous and cause wrong classifications upon deployment.

\medskip

Overall, this use case has demonstrated how our method can help detect differences between datasets and present them interpretably to the user. We see such functionality as crucial for dataset consolidation, a key step towards training better models and designing better evaluation benchmarks.

\section{Conclusion}
\label{sec:conclusion}
This work has addressed how to gain insight into real-world distribution shifts and transport phenomena. To this end, we introduced \mname{}, a novel framework to explain Wasserstein distances---a popular tool for comparing distributions---by attributing them to distinct data components, such as individual features or instances. To the best of our knowledge, no other works have focused on attributing Wasserstein distances so far, and our work thereby fills a gap in the existing XAI literature. Our method is theoretically supported, computationally efficient, and shows broad applicability by flexibly adapting to various Wasserstein specifications and dataset types. We demonstrated our method's high accuracy and utility through a comprehensive evaluation against a diverse set of baselines.
Moreover, we showed how \mname{} can aid in identifying relevant features for aligning domains, thereby improving classifier robustness.

Our method is also designed to be interoperable with subspace analysis techniques, enabling the extension \wdsa{}, which disentangles Wasserstein models into multiple transport components. The usefulness of \wdsa{}'s disentangling capabilities was demonstrated in two realistic use cases: gaining insights into an abalone aging phenomenon, and uncovering subtle differences between two popular face datasets.

It is important to emphasize that our method does not directly describe the ground-truth transport phenomena but provides an interpretation of the Wasserstein distance used to model them. Consequently, any limitations of the Wasserstein model---such as its sensitivity to noise or its preference for least action---will be reflected in the resulting explanation. We therefore position our method as an interactive tool for practitioners: it can aid in validating a given transport model, choosing its hyperparameters, or developing more informed Wasserstein-based models. This, in turn, enables a more systematic and insightful analysis of the observed transport phenomena or dataset shift.

While we have focused in this work on classical optimal transport formulations, more advanced formulations that address their limitations (cf.~\cite{alvarez-melisStructuredOptimalTransport2018, chizatScalingAlgorithmsUnbalanced2018, chapelPartialOptimalTranport2020, mukherjeeOutlierrobustOptimalTransport2021}) should be explored in future work. This may include extending \mname{} to Gromov-Wasserstein distances~\cite{memoliGromovWassersteinDistances2011} (and its extensions~\cite{titouanCOoptimalTransport2020}), which allow transport across unregistered spaces, and that can also be expressed as a hierarchy of distances and norms. Another formulation is the `Sliced Wasserstein distance'~\cite{bonneelSlicedRadonWasserstein2015}, which approximates Wasserstein distances through multiple estimations on random 1-dimensional subspaces in order to scale up to large datasets. As our method, specifically \wdsa{}, also operates on subspaces, it may readily include some of the building blocks necessary to explain these large-scale transport models. Furthermore, while our work has demonstrated the effectiveness of \mname{} at gaining insight into transport phenomena, achieving finer characterizations by adapting it to more refined transport models (e.g.~accounting for temporal dynamics and causality~\cite{bartlWassersteinSpaceStochastic2024}) appears to be another interesting future direction.

\section*{Acknowledgment}
\noindent This work was in part supported by the Federal
German Ministry for Education and Research (BMBF) under Grants
BIFOLD24B, BIFOLD25B, 01IS18037A, 01IS18025A, and 01IS24087C. We thank K.-R.\ M\"uller for the valuable comments on the manuscript.



\begin{thebibliography}{10}
\providecommand{\url}[1]{#1}
\csname url@samestyle\endcsname
\providecommand{\newblock}{\relax}
\providecommand{\bibinfo}[2]{#2}
\providecommand{\BIBentrySTDinterwordspacing}{\spaceskip=0pt\relax}
\providecommand{\BIBentryALTinterwordstretchfactor}{4}
\providecommand{\BIBentryALTinterwordspacing}{\spaceskip=\fontdimen2\font plus
\BIBentryALTinterwordstretchfactor\fontdimen3\font minus \fontdimen4\font\relax}
\providecommand{\BIBforeignlanguage}[2]{{%
\expandafter\ifx\csname l@#1\endcsname\relax
\typeout{** WARNING: IEEEtran.bst: No hyphenation pattern has been}%
\typeout{** loaded for the language `#1'. Using the pattern for}%
\typeout{** the default language instead.}%
\else
\language=\csname l@#1\endcsname
\fi
#2}}
\providecommand{\BIBdecl}{\relax}
\BIBdecl

\bibitem{kleinMappingCellsTime2025}
D.~Klein, G.~Palla, M.~Lange, M.~Klein, Z.~Piran, M.~Gander, L.~{Meng-Papaxanthos}, M.~Sterr, L.~Saber, C.~Jing, A.~{Bastidas-Ponce}, P.~Cota, M.~{Tarquis-Medina}, S.~Parikh, I.~Gold, H.~Lickert, M.~Bakhti, M.~Nitzan, M.~Cuturi, and F.~J. Theis, ``Mapping cells through time and space with {Moscot},'' \emph{Nature}, vol. 638, no. 8052, pp. 1065--1075, Feb. 2025.

\bibitem{villaniOptimalTransportOld2008}
C.~Villani, \emph{Optimal Transport: {{Old}} and New}, ser. Grundlehren Der Mathematischen Wissenschaften.\hskip 1em plus 0.5em minus 0.4em\relax Springer Berlin Heidelberg, 2008.

\bibitem{peyreComputationalOptimalTransport2020}
G.~Peyr{\'e} and M.~Cuturi, ``Computational {{Optimal Transport}},'' Mar. 2020.

\bibitem{montesumaRecentAdvancesOptimal2025}
E.~F. Montesuma, F.~M.~N. Mboula, and A.~Souloumiac, ``Recent advances in optimal transport for machine learning,'' \emph{IEEE Transactions on Pattern Analysis and Machine Intelligence}, vol.~47, no.~2, pp. 1161--1180, 2025.

\bibitem{cuturiSinkhornDistancesLightspeed2013}
M.~Cuturi, ``Sinkhorn {{Distances}}: {{Lightspeed Computation}} of {{Optimal Transport}},'' in \emph{Advances in {{Neural Information Processing Systems}}}, vol.~26.\hskip 1em plus 0.5em minus 0.4em\relax Curran Associates, Inc., 2013, pp. 2292--2300.

\bibitem{kulinskiExplainingDistributionShifts2023}
S.~Kulinski and D.~I. Inouye, ``Towards explaining distribution shifts,'' in \emph{{ICML}}, ser. Proceedings of Machine Learning Research, vol. 202.\hskip 1em plus 0.5em minus 0.4em\relax {PMLR}, 2023, pp. 17\,931--17\,952.

\bibitem{montavonMethodsInterpretingUnderstanding2018}
G.~Montavon, W.~Samek, and K.-R. M{\"{u}}ller, ``Methods for interpreting and understanding deep neural networks,'' \emph{Digital Signal Processing}, vol.~73, pp. 1--15, 2018.

\bibitem{gunningDARPAsExplainableArtificial2019}
D.~Gunning and D.~W. Aha, ``{{DARPA}}'s explainable artificial intelligence ({{XAI}}) program,'' \emph{{AI} {Magazine}}, vol.~40, no.~2, pp. 44--58, 2019.

\bibitem{samekExplainingDeepNeural2021}
W.~Samek, G.~Montavon, S.~Lapuschkin, C.~J. Anders, and K.-R. M{\"u}ller, ``Explaining {{Deep Neural Networks}} and {{Beyond}}: {{A Review}} of {{Methods}} and {{Applications}},'' \emph{Proceedings of the IEEE}, vol. 109, no.~3, pp. 247--278, Mar. 2021.

\bibitem{barredoarrietaExplainableArtificialIntelligence2020}
A.~Barredo~Arrieta, N.~{D{\'i}az-Rodr{\'i}guez}, J.~Del~Ser, A.~Bennetot, S.~Tabik, A.~Barbado, S.~Garcia, S.~{Gil-Lopez}, D.~Molina, R.~Benjamins, R.~Chatila, and F.~Herrera, ``Explainable {{Artificial Intelligence}} ({{XAI}}): {{Concepts}}, taxonomies, opportunities and challenges toward responsible {{AI}},'' \emph{Information Fusion}, vol.~58, pp. 82--115, Jun. 2020.

\bibitem{lapuschkinUnmaskingCleverHans2019}
S.~Lapuschkin, S.~W{\"a}ldchen, A.~Binder, G.~Montavon, W.~Samek, and K.-R. M{\"u}ller, ``Unmasking {{Clever Hans}} predictors and assessing what machines really learn,'' \emph{Nature Communications}, vol.~10, no.~1, p. 1096, Mar. 2019.

\bibitem{degraveAIRadiographicCOVID192021}
A.~J. DeGrave, J.~D. Janizek, and S.-I. Lee, ``{{AI}} for radiographic {{COVID-19}} detection selects shortcuts over signal,'' \emph{Nature Machine Intelligence}, vol.~3, no.~7, pp. 610--619, Jul. 2021.

\bibitem{kauffmannExplainableAIReveals2025}
J.~Kauffmann, J.~Dippel, L.~Ruff, W.~Samek, K.-R. M{\"u}ller, and G.~Montavon, ``Explainable {{AI}} reveals {{Clever Hans}} effects in unsupervised learning models,'' \emph{Nature Machine Intelligence}, vol.~7, no.~3, pp. 412--422, Mar. 2025.

\bibitem{binderMorphologicalMolecularBreast2021}
A.~Binder, M.~Bockmayr, M.~H{\"{a}}gele, S.~Wienert, D.~Heim, K.~Hellweg, M.~Ishii, A.~Stenzinger, A.~Hocke, C.~Denkert, K.-R. M{\"{u}}ller, and F.~Klauschen, ``Morphological and molecular breast cancer profiling through explainable machine learning,'' \emph{Nature Machine Intelligence}, vol.~3, no.~4, pp. 355--366, 2021.

\bibitem{klauschenExplainableArtificialIntelligence2024}
F.~Klauschen, J.~Dippel, P.~Keyl, P.~Jurmeister, M.~Bockmayr, A.~Mock, O.~Buchstab, M.~Alber, L.~Ruff, G.~Montavon, and K.-R. M\"{u}ller, ``Toward explainable artificial intelligence for precision pathology,'' \emph{Annual Review of Pathology: Mechanisms of Disease}, vol.~19, no.~1, pp. 541--570, 2024.

\bibitem{zednikScientificExplorationExplainable2022}
C.~Zednik and H.~Boelsen, ``Scientific {{Exploration}} and {{Explainable Artificial Intelligence}},'' \emph{Minds and Machines}, vol.~32, no.~1, pp. 219--239, Mar. 2022.

\bibitem{roscherExplainableMachineLearning2020}
R.~Roscher, B.~Bohn, M.~F. Duarte, and J.~Garcke, ``Explainable {{Machine Learning}} for {{Scientific Insights}} and {{Discoveries}},'' \emph{IEEE Access}, vol.~8, pp. 42\,200--42\,216, 2020.

\bibitem{strumbeljExplainingPredictionModels2014}
E.~{\v S}trumbelj and I.~Kononenko, ``Explaining prediction models and individual predictions with feature contributions,'' \emph{Knowledge and Information Systems}, vol.~41, no.~3, pp. 647--665, Dec. 2014.

\bibitem{lundbergUnifiedApproachInterpreting2017}
S.~M. Lundberg and S.-I. Lee, ``A {{Unified Approach}} to {{Interpreting Model Predictions}},'' in \emph{Advances in {{Neural Information Processing Systems}}}, vol.~30.\hskip 1em plus 0.5em minus 0.4em\relax Curran Associates, Inc., 2017.

\bibitem{bachPixelwiseExplanationsNonlinear2015}
S.~Bach, A.~Binder, G.~Montavon, F.~Klauschen, K.-R. M{\"u}ller, and W.~Samek, ``On pixel-wise explanations for non-linear classifier decisions by layer-wise relevance propagation,'' \emph{PLOS ONE}, vol.~10, no.~7, pp. 1--46, Jul. 2015.

\bibitem{montavonLayerWiseRelevancePropagation2019}
G.~Montavon, A.~Binder, S.~Lapuschkin, W.~Samek, and K.-R. M{\"{u}}ller, ``Layer-wise relevance propagation: An overview,'' in \emph{Explainable {AI}}, ser. Lecture Notes in Computer Science.\hskip 1em plus 0.5em minus 0.4em\relax Springer, 2019, vol. 11700, pp. 193--209.

\bibitem{kimInterpretabilityFeatureAttribution2018}
B.~Kim, M.~Wattenberg, J.~Gilmer, C.~J. Cai, J.~Wexler, F.~B. Vi{\'{e}}gas, and R.~Sayres, ``Interpretability beyond feature attribution: Quantitative testing with concept activation vectors {(TCAV)},'' in \emph{{ICML}}, ser. Proceedings of Machine Learning Research, vol.~80.\hskip 1em plus 0.5em minus 0.4em\relax {PMLR}, 2018, pp. 2673--2682.

\bibitem{achtibatAttributionMapsHumanunderstandable2023}
R.~Achtibat, M.~Dreyer, I.~Eisenbraun, S.~Bosse, T.~Wiegand, W.~Samek, and S.~Lapuschkin, ``From attribution maps to human-understandable explanations through {{Concept Relevance Propagation}},'' \emph{Nature Machine Intelligence}, vol.~5, no.~9, pp. 1006--1019, Sep. 2023.

\bibitem{chormaiDisentangledExplanationsNeural2024}
P.~Chormai, J.~Herrmann, K.-R. M{\"u}ller, and G.~Montavon, ``Disentangled explanations of neural network predictions by finding relevant subspaces,'' \emph{IEEE Transactions on Pattern Analysis and Machine Intelligence}, vol.~46, no.~11, pp. 7283--7299, 2024.

\bibitem{quionero-candelaDatasetShiftMachine2009}
J.~{Quionero-Candela}, M.~Sugiyama, A.~Schwaighofer, and N.~D. Lawrence, \emph{Dataset Shift in Machine Learning}.\hskip 1em plus 0.5em minus 0.4em\relax The MIT Press, 2009.

\bibitem{kohWILDSBenchmarkIntheWild2021}
P.~W. Koh, S.~Sagawa, H.~Marklund, S.~M. Xie, M.~Zhang, A.~Balsubramani, W.~Hu, M.~Yasunaga, R.~L. Phillips, I.~Gao, T.~Lee, E.~David, I.~Stavness, W.~Guo, B.~Earnshaw, I.~S. Haque, S.~M. Beery, J.~Leskovec, A.~Kundaje, E.~Pierson, S.~Levine, C.~Finn, and P.~Liang, ``{WILDS:} {A} benchmark of in-the-wild distribution shifts,'' in \emph{{ICML}}, ser. Proceedings of Machine Learning Research, vol. 139.\hskip 1em plus 0.5em minus 0.4em\relax {PMLR}, 2021, pp. 5637--5664.

\bibitem{courtyOptimalTransportDomain2017}
N.~Courty, R.~Flamary, D.~Tuia, and A.~Rakotomamonjy, ``Optimal {{Transport}} for {{Domain Adaptation}},'' \emph{IEEE Transactions on Pattern Analysis and Machine Intelligence}, vol.~39, no.~9, pp. 1853--1865, Sep. 2017.

\bibitem{budhathokiWhyDidDistribution2021}
K.~Budhathoki, D.~Janzing, P.~Bl{\"{o}}baum, and H.~Ng, ``Why did the distribution change?'' in \emph{{AISTATS}}, ser. Proceedings of Machine Learning Research, vol. 130.\hskip 1em plus 0.5em minus 0.4em\relax {PMLR}, 2021, pp. 1666--1674.

\bibitem{babbarWhatDifferentThese2025}
V.~Babbar, Z.~Guo, and C.~Rudin, ````{{What}} is different between these datasets?'' {A} framework for explaining data distribution shifts,'' \emph{Journal of Machine Learning Research}, vol.~26, no. 180, pp. 1--64, 2025.

\bibitem{donnellyRashomonImportanceDistribution2023}
J.~Donnelly, S.~Katta, C.~Rudin, and E.~P. Browne, ``The {Rashomon Importance Distribution}: {{Getting RID}} of unstable, single model-based variable importance,'' in \emph{Advances in {{Neural Information Processing Systems}}}, vol.~36.\hskip 1em plus 0.5em minus 0.4em\relax Curran Associates, Inc., 2023, pp. 6267--6279.

\bibitem{hulkundInterpretableDistributionShift2022}
N.~Hulkund, N.~Fusi, J.~W. Vaughan, and D.~{Alvarez-Melis}, ``Interpretable {{Distribution Shift Detection}} using {{Optimal Transport}},'' Aug. 2022.

\bibitem{gautheronFeatureSelectionUnsupervised2018}
L.~Gautheron, I.~Redko, and C.~Lartizien, ``Feature selection for unsupervised domain adaptation using optimal transport,'' in \emph{{{ECML}}/{{PKDD}} (2)}, ser. Lecture Notes in Computer Science, vol. 11052.\hskip 1em plus 0.5em minus 0.4em\relax Springer, 2018, pp. 759--776.

\bibitem{sundararajanAxiomaticAttributionDeep2017}
M.~Sundararajan, A.~Taly, and Q.~Yan, ``Axiomatic attribution for deep networks,'' in \emph{{ICML}}, ser. Proceedings of Machine Learning Research, vol.~70.\hskip 1em plus 0.5em minus 0.4em\relax {PMLR}, 2017, pp. 3319--3328.

\bibitem{linMakingTransportMore2021}
C.~Lin, M.~Azabou, and E.~L. Dyer, ``Making transport more robust and interpretable by moving data through a small number of anchor points,'' in \emph{{ICML}}, ser. Proceedings of Machine Learning Research, vol. 139.\hskip 1em plus 0.5em minus 0.4em\relax {PMLR}, 2021, pp. 6631--6641.

\bibitem{cuturiMongeBregmanOccam2023}
M.~Cuturi, M.~Klein, and P.~Ablin, ``{Monge}, {Bregman} and {Occam}: Interpretable optimal transport in high-dimensions with feature-sparse maps,'' in \emph{{ICML}}, ser. Proceedings of Machine Learning Research, vol. 202.\hskip 1em plus 0.5em minus 0.4em\relax {PMLR}, 2023, pp. 6671--6682.

\bibitem{zhangWhyDidModel2023}
H.~Zhang, H.~Singh, M.~Ghassemi, and S.~Joshi, ````{Why} did the model fail?'': Attributing model performance changes to distribution shifts,'' in \emph{{ICML}}, ser. Proceedings of Machine Learning Research, vol. 202.\hskip 1em plus 0.5em minus 0.4em\relax {PMLR}, 2023, pp. 41\,550--41\,578.

\bibitem{deckerExplanatoryModelMonitoring2024}
T.~Decker, A.~Koebler, M.~Lebacher, I.~Thon, V.~Tresp, and F.~Buettner, ``Explanatory model monitoring to understand the effects of feature shifts on performance,'' in \emph{{KDD}}.\hskip 1em plus 0.5em minus 0.4em\relax {ACM}, 2024, pp. 550--561.

\bibitem{liuNeedLanguageDescribing2023}
J.~Liu, T.~Wang, P.~Cui, and H.~Namkoong, ``On the {{Need}} for a {{Language Describing Distribution Shifts}}: {{Illustrations}} on {{Tabular Datasets}},'' in \emph{Advances in {{Neural Information Processing Systems}}}, vol.~36.\hskip 1em plus 0.5em minus 0.4em\relax Curran Associates, Inc., Nov. 2023, pp. 51\,371--51\,408.

\bibitem{gardnerBenchmarkingDistributionShift2023}
J.~Gardner, Z.~Popovic, and L.~Schmidt, ``Benchmarking distribution shift in tabular data with {{TableShift}},'' in \emph{Advances in {{Neural Information Processing Systems}}}, vol.~36.\hskip 1em plus 0.5em minus 0.4em\relax Curran Associates, Inc., 2023, pp. 53\,385--53\,432.

\bibitem{montavonExplainingPredictionsUnsupervised2022}
G.~Montavon, J.~R. Kauffmann, W.~Samek, and K.-R. M{\"{u}}ller, ``Explaining the predictions of unsupervised learning models,'' in \emph{xxAI@ICML}, ser. Lecture Notes in Computer Science.\hskip 1em plus 0.5em minus 0.4em\relax Springer, 2022, vol. 13200, pp. 117--138.

\bibitem{kauffmannClusteringClusterExplanations2024}
J.~Kauffmann, M.~Esders, L.~Ruff, G.~Montavon, W.~Samek, and K.-R. M{\"u}ller, ``From {{Clustering}} to {{Cluster Explanations}} via {{Neural Networks}},'' \emph{IEEE Transactions on Neural Networks and Learning Systems}, vol.~35, no.~2, pp. 1926--1940, Feb. 2024.

\bibitem{youDistributionalCounterfactualExplanations2025}
L.~You, L.~Cao, M.~Nilsson, B.~Zhao, and L.~Lei, ``Distributional counterfactual explanations with optimal transport,'' in \emph{{{AISTATS}}}, ser. Proceedings of Machine Learning Research, vol. 258.\hskip 1em plus 0.5em minus 0.4em\relax PMLR, 2025, pp. 1135--1143.

\bibitem{serrurierExplainableProperties1Lipschitz2023}
M.~Serrurier, F.~Mamalet, T.~Fel, L.~B{\'e}thune, and T.~Boissin, ``On the explainable properties of 1-{{Lipschitz Neural Networks}}: {{An Optimal Transport Perspective}},'' in \emph{Advances in {{Neural Information Processing Systems}}}, vol.~36.\hskip 1em plus 0.5em minus 0.4em\relax Curran Associates, Inc., Nov. 2023, pp. 54\,645--54\,682.

\bibitem{hillAxiomaticExplainerGlobalness2025}
D.~Hill, J.~Bone, A.~Masoomi, M.~Torop, and J.~Dy, ``Axiomatic explainer globalness via optimal transport,'' in \emph{{AISTATS}}, 2025.

\bibitem{selvarajuGradCAMVisualExplanations2020}
R.~R. Selvaraju, M.~Cogswell, A.~Das, R.~Vedantam, D.~Parikh, and D.~Batra, ``{Grad-CAM}: Visual explanations from deep networks via gradient-based localization,'' \emph{Int. J. Comput. Vis.}, vol. 128, no.~2, pp. 336--359, 2020.

\bibitem{aliXAITransformersBetter2022}
A.~Ali, T.~Schnake, O.~Eberle, G.~Montavon, K.-R. M{\"{u}}ller, and L.~Wolf, ``{XAI} for transformers: Better explanations through conservative propagation,'' in \emph{{ICML}}, ser. Proceedings of Machine Learning Research, vol. 162.\hskip 1em plus 0.5em minus 0.4em\relax {PMLR}, 2022, pp. 435--451.

\bibitem{schnakeHigherorderExplanationsGraph2022}
T.~Schnake, O.~Eberle, J.~Lederer, S.~Nakajima, K.~T. Sch{\"u}tt, K.-R. M{\"u}ller, and G.~Montavon, ``Higher-order explanations of graph neural networks via relevant walks,'' \emph{IEEE Transactions on Pattern Analysis and Machine Intelligence}, vol.~44, no.~11, pp. 7581--7596, 2022.

\bibitem{bluecherDecouplingPixelFlipping2024}
S.~Bluecher, J.~Vielhaben, and N.~Strodthoff, ``Decoupling pixel flipping and occlusion strategy for consistent {XAI} benchmarks,'' \emph{Transactions on Machine Learning Research}, vol. 2024, 2024.

\bibitem{samekEvaluatingVisualizationWhat2017}
W.~Samek, A.~Binder, G.~Montavon, S.~Lapuschkin, and K.-R. M{\"u}ller, ``Evaluating the {{Visualization}} of {{What}} a {{Deep Neural Network Has Learned}},'' \emph{IEEE Transactions on Neural Networks and Learning Systems}, vol.~28, no.~11, pp. 2660--2673, Nov. 2017.

\bibitem{filiotDistillingFoundationModels2025}
A.~Filiot, N.~Dop, O.~Tchita, A.~Riou, R.~Dubois, T.~Peeters, D.~Valter, M.~Scalbert, C.~Saillard, G.~Robin, and A.~Olivier, ``Distilling foundation models for robust and efficient models in digital pathology,'' 2025.

\bibitem{cigaSelfSupervisedContrastive2022}
O.~Ciga, T.~Xu, and A.~L. Martel, ``Self supervised contrastive learning for digital histopathology,'' \emph{Machine Learning with Applications}, vol.~7, p. 100198, Mar. 2022.

\bibitem{filiotScalingSelfSupervisedLearning2024}
A.~Filiot, R.~Ghermi, A.~Olivier, P.~Jacob, L.~Fidon, A.~Camara, A.~M. Kain, C.~Saillard, and J.-B. Schiratti, ``Scaling {{Self-Supervised Learning}} for {{Histopathology}} with {{Masked Image Modeling}},'' p. 2023.07.21.23292757, Dec. 2024.

\bibitem{chenGeneralpurposeFoundationModel2024}
R.~J. Chen, T.~Ding, M.~Y. Lu, D.~F.~K. Williamson, G.~Jaume, A.~H. Song, B.~Chen, A.~Zhang, D.~Shao, M.~Shaban, M.~Williams, L.~Oldenburg, L.~L. Weishaupt, J.~J. Wang, A.~Vaidya, L.~P. Le, G.~Gerber, S.~Sahai, W.~Williams, and F.~Mahmood, ``Towards a general-purpose foundation model for computational pathology,'' \emph{Nature Medicine}, vol.~30, no.~3, pp. 850--862, Mar. 2024.

\bibitem{haufeInterpretationWeightVectors2014}
S.~Haufe, F.~Meinecke, K.~G{\"o}rgen, S.~D{\"a}hne, J.-D. Haynes, B.~Blankertz, and F.~Bie{\ss}mann, ``On the interpretation of weight vectors of linear models in multivariate neuroimaging,'' \emph{NeuroImage}, vol.~87, pp. 96--110, 2014.

\bibitem{komenRobustFoundationModels2025}
J.~K{\"o}men, E.~D. {de Jong}, J.~Hense, H.~Marienwald, J.~Dippel, P.~Naumann, E.~Marcus, L.~Ruff, M.~Alber, J.~Teuwen, F.~Klauschen, and K.-R. M{\"u}ller, ``Towards robust foundation models for digital pathology,'' 2025.

\bibitem{braunRelevantDimensionsKernel2008}
M.~L. Braun, J.~M. Buhmann, and K.-R. M{\"u}ller, ``On relevant dimensions in kernel feature spaces,'' \emph{{Journal of Machine Learning Research}}, vol.~9, pp. 1875--1908, 2008.

\bibitem{gongGeodesicFlowKernel2012}
B.~Gong, Y.~Shi, F.~Sha, and K.~Grauman, ``Geodesic flow kernel for unsupervised domain adaptation,'' in \emph{{{CVPR}}}.\hskip 1em plus 0.5em minus 0.4em\relax {IEEE} Computer Society, 2012, pp. 2066--2073.

\bibitem{nashAbalone1995}
W.~Nash, T.~Sellers, S.~Talbot, A.~Cawthorn, and W.~Ford, ``Abalone,'' \url{https://doi.org/10.24432/C55C7W}, Nov. 1995.

\bibitem{alvarez-melisGeometricDatasetDistances2020}
D.~Alvarez{-}Melis and N.~Fusi, ``Geometric dataset distances via optimal transport,'' in \emph{Advances in {{Neural Information Processing Systems}}}, vol.~33.\hskip 1em plus 0.5em minus 0.4em\relax Curran Associates, Inc., 2020, pp. 21\,428--21\,439.

\bibitem{liuDeepLearningFace2015}
Z.~Liu, P.~Luo, X.~Wang, and X.~Tang, ``Deep learning face attributes in the wild,'' in \emph{{ICCV}}.\hskip 1em plus 0.5em minus 0.4em\relax {IEEE} Computer Society, 2015, pp. 3730--3738.

\bibitem{huangLabeledFacesWild2007}
G.~B. Huang, M.~Ramesh, T.~Berg, and E.~{Learned-Miller}, ``Labeled faces in the wild: A database for studying face recognition in unconstrained environments,'' University of Massachusetts, Amherst, Tech. Rep. 07-49, Oct. 2007.

\bibitem{radfordLearningTransferableVisual2021}
A.~Radford, J.~W. Kim, C.~Hallacy, A.~Ramesh, G.~Goh, S.~Agarwal, G.~Sastry, A.~Askell, P.~Mishkin, J.~Clark, G.~Krueger, and I.~Sutskever, ``Learning transferable visual models from natural language supervision,'' in \emph{{ICML}}, ser. Proceedings of Machine Learning Research, vol. 139.\hskip 1em plus 0.5em minus 0.4em\relax {PMLR}, 2021, pp. 8748--8763.

\bibitem{oikarinenCLIPdissectAutomaticDescription2023}
T.~P. Oikarinen and T.~Weng, ``{CLIP-Dissect}: Automatic description of neuron representations in deep vision networks,'' in \emph{{ICLR}}.\hskip 1em plus 0.5em minus 0.4em\relax OpenReview.net, 2023.

\bibitem{alvarez-melisStructuredOptimalTransport2018}
D.~Alvarez{-}Melis, T.~S. Jaakkola, and S.~Jegelka, ``Structured optimal transport,'' in \emph{{AISTATS}}, ser. Proceedings of Machine Learning Research, vol.~84.\hskip 1em plus 0.5em minus 0.4em\relax {PMLR}, 2018, pp. 1771--1780.

\bibitem{chizatScalingAlgorithmsUnbalanced2018}
L.~Chizat, G.~Peyr{\'e}, B.~Schmitzer, and F.-X. Vialard, ``Scaling algorithms for unbalanced optimal transport problems,'' \emph{Math. Comput.}, vol.~87, no. 314, pp. 2563--2609, 2018.

\bibitem{chapelPartialOptimalTranport2020}
L.~Chapel, M.~Z. Alaya, and G.~Gasso, ``Partial {{Optimal Tranport}} with applications on {{Positive-Unlabeled Learning}},'' in \emph{Advances in {{Neural Information Processing Systems}}}, vol.~33.\hskip 1em plus 0.5em minus 0.4em\relax Curran Associates, Inc., 2020, pp. 2903--2913.

\bibitem{mukherjeeOutlierrobustOptimalTransport2021}
D.~Mukherjee, A.~Guha, J.~M. Solomon, Y.~Sun, and M.~Yurochkin, ``Outlier-robust optimal transport,'' in \emph{{ICML}}, ser. Proceedings of Machine Learning Research, vol. 139.\hskip 1em plus 0.5em minus 0.4em\relax {PMLR}, 2021, pp. 7850--7860.

\bibitem{memoliGromovWassersteinDistances2011}
F.~M{\'e}moli, ``Gromov--{{Wasserstein Distances}} and the {{Metric Approach}} to {{Object Matching}},'' \emph{Foundations of Computational Mathematics}, vol.~11, no.~4, pp. 417--487, Aug. 2011.

\bibitem{titouanCOoptimalTransport2020}
V.~Titouan, I.~Redko, R.~Flamary, and N.~Courty, ``{{CO-optimal} Transport},'' in \emph{Advances in {{Neural Information Processing Systems}}}, vol.~33.\hskip 1em plus 0.5em minus 0.4em\relax Curran Associates, Inc., 2020, pp. 17\,559--17\,570.

\bibitem{bonneelSlicedRadonWasserstein2015}
N.~Bonneel, J.~Rabin, G.~Peyr{\'e}, and H.~Pfister, ``{Sliced and Radon Wasserstein Barycenters of Measures},'' \emph{Journal of Mathematical Imaging and Vision}, vol.~51, no.~1, pp. 22--45, 2015.

\bibitem{bartlWassersteinSpaceStochastic2024}
D.~Bartl, M.~Beiglb{\"o}ck, and G.~Pammer, ``The {{Wasserstein}} space of stochastic processes,'' \emph{Journal of the European Mathematical Society}, Dec. 2024.

\end{thebibliography}
\end{document}